\documentclass{article}

\usepackage{PRIMEarxiv}
\usepackage{amsmath}
\usepackage[utf8]{inputenc} % allow utf-8 input
\usepackage[T1,T2A]{fontenc}
\usepackage{hyperref}       % hyperlinks
\usepackage{url}            % simple URL typesetting
\usepackage{booktabs}       % professional-quality tables
\usepackage{amsfonts}       % blackboard math symbols
\usepackage{nicefrac}       % compact symbols for 1/2, etc.
\usepackage{microtype}      % microtypography
\usepackage{lipsum}
\usepackage{fancyhdr}       % header
\usepackage{graphicx}       % graphics
\graphicspath{{media/}}     % organize your images and other figures under media/ folder
\usepackage{multirow}
\title{A Learnable Multi-views Contrastive Framework with Reconstruction Discrepancy for Medical Time-Series
}

\author{
  Yifan Wang\thanks{These authors contributed equally.} \\
  School of Medicine, \\
  The Chinese University of Hong Kong, Shenzhen \\
  \texttt{(yifanwang13@link.cuhk.edu.cn)} \\
  \and
  Hongfeng Ai\footnotemark[1] \\
  Digital Department, \\
  China Unicom Global Co., Ltd \\
  \texttt{(aihf@chinaunicom.cn)} \\
  \and
  Ruiqi Li\footnotemark[1] \\
  University of the Chinese Academy of Sciences \\
  \texttt{(liruiqi1@sia.cn)} \\
  \and
  Maowei Jiang\footnotemark[1] \\
  Shenzhen International Graduate School, \\
  Tsinghua University \\
  \texttt{(jiangmaowei@sia.cn)} \\
  \and
  Cheng Jiang\thanks{Corresponding author.} \\
  School of Medicine, \\
  The Chinese University of Hong Kong, Shenzhen \\
  \texttt{jiangcheng@cuhk.edu.cn} \\
  \and
  Chenzhong Li\footnotemark[2] \\
  School of Medicine, \\
  The Chinese University of Hong Kong, Shenzhen \\
  \texttt{lichenzhong@cuhk.edu.cn} \\
}

\begin{document}
\maketitle

\begin{abstract}
In medical time series disease diagnosis, two key challenges are identified. First, the high annotation cost of medical data leads to overfitting in models trained on label-limited, single-center datasets. To address this, we propose incorporating external data from related tasks and leveraging AE-GAN to extract prior knowledge, providing valuable references for downstream tasks. Second, many existing studies employ contrastive learning to derive more generalized medical sequence representations for diagnostic tasks, usually relying on manually designed diverse positive and negative sample pairs. However, these approaches are complex, lack generalizability, and fail to adaptively capture disease-specific features across different conditions. To overcome this, we introduce LMCF (Learnable Multi-views Contrastive Framework), a framework that integrates a multi-head attention mechanism and adaptively learns representations from different views through inter-view and intra-view contrastive learning strategies. Additionally, the pre-trained AE-GAN is used to reconstruct discrepancies in the target data as disease probabilities, which are then integrated into the contrastive learning process. Experiments on three target datasets demonstrate that our method consistently outperforms other seven baselines, highlighting its significant impact on healthcare applications such as the diagnosis of myocardial infarction, Alzheimer’s disease, and Parkinson’s disease. We release the source code at xxxxx.
\end{abstract}

% keywords can be removed

\section{Introduction}
Medical time series data plays a critical role in various healthcare applications, ranging from disease diagnosis to patient monitoring, including electroencephalography (EEG), electrocardiography (ECG), blood pressure monitoring, respiratory rate tracking, oxygen saturation (SpO2) measurements, and more.However, the inherent complexity and variety of these signals make the process of labeling not only time-consuming but also requiring expert knowledge, typically from clinicians, to manually annotate the data. This limitation makes it particularly difficult to apply deep learning techniques, which generally rely on large amounts of labeled data to achieve optimal performance.To overcome the challenge of limited labeled data in time series, self-supervised contrastive learning has emerged as a compelling and effective approach. Recent works \cite{liu2024timesurl,choi2024multi,liang2024factorized} have demonstrated the effectiveness of contrastive learning in medical time series across various applications. Different from computer vision where each image is a sample and the positive or negative sample is certainly at the image-level, the data in medical time series is organized hierarchically \cite{brunekreef2024kandinsky}. For instance, TS2Vec \cite{yue2022ts2vec} hierarchically distinguishes positive and negative samples across both instance-wise and temporal dimensions. CLOCS \cite{kiyasseh2021clocs} divides the entire patient's data into two levels: Temporal Invariance and Spatial Invariance. COMET \cite{wang2024contrast} divides medical data into four levels: Patient, Trial, Sample, and Observation, to capture information at different granularities.However, in the context of using medical time series data for disease diagnosis, two key issues have emerged based on prior work. The first issue is the reliance on single-center datasets, which introduce regional biases and limit model generalizability.These datasets often have small sample sizes and lack diversity, making it difficult to apply models to other healthcare environments\cite{chang2023mining}.The sequence patterns of normal patients in external data encompass rich knowledge that enhances the diversity and representativeness of the dataset, mitigating the biases inherent in single-center data. The second issue is that the primary challenge of contrastive learning lies in constructing positive and negative samples \cite{liu2021self}. Many existing studies attempt to address this by manually designing diverse contrastive dimensions to capture temporal information at varying granularities, leveraging self-supervised learning to derive representations that improve the generalization performance of downstream label-limited tasks. However, unlike images or texts, medical time series data exhibit more subtle and complex variations, making it inherently difficult to manually define comprehensive and accurate positive and negative sample pairs \cite{liu2023self}. This process is not only labor-intensive but also constrained in its ability to fully capture the intricate characteristics of medical time series, ultimately limiting its effectiveness in real-world applications.In this paper, we propose a novel approach LMCRD to address the aforementioned challenges. The key contributions of this work can be summarized as follows:
\begin{itemize}
\item We leverage external data and AE-GAN to extract prior knowledge, enhancing model generalization through cross-center knowledge transfer. Practically, AE-GAN reconstructs target data discrepancies as disease probabilities, which are integrated into the contrastive learning process to improve the learning of the target representation.
\item We propose the LMCF, which integrates a multi-head attention mechanism (MHA) and adaptive contrastive learning strategies from both inter-view and intra-view considerations. This framework enables the model to learn robust representations from diverse views while mitigating the reliance on manually designed positive and negative sample pairs. This contribution enhances the flexibility and efficiency of representation learning.
\item 
We demonstrate that our method achieves state-of-the-art (SOTA) performance across three critical datasets—myocardial infarction, Alzheimer’s disease, and Parkinson’s disease. Particularly noteworthy is that even in an extreme scenario where only 10\% of the data are labeled, our method continues to deliver leading performance.
\end{itemize}
\section{Related work}
\subsection{Multiple centers medical time series}
In recent years, research on optimizing models using multi-center medical datasets has garnered increasing attention \cite{wang2021deep, linardos2022federated, asadi2024learning}. These researches are motivated by the fact that deep learning models often overfit when trained solely on data from a single medical center. Especially for the medical time series datasets, such as those involving EEG and ECG, often lack sufficient scale and comprehensive annotations. While domain adaptation \cite{ganin2016domain, asadi2024learning} is proposed for mapping features from the external domain to a similar feature space of the target domain and then using the classifier from the data-rich source domain to classify data in the data-scarce target domain. This method has limitations in the context of medical time series. Specifically, the difference in the scale of data between the external and target domains is not always significant. As a result, leveraging external data to enhance the performance of models trained on the target domain has become a preferred strategy. Therefore, transferring knowledge across centers to improve model performance on the target center's data distribution is an important direction for our current work.
\subsection{Contrastive learning for time series}
Self-supervised learning leverages unlabeled data to pretrain models, enabling them to capture temporal dependencies and meaningful features without relying on costly annotations. Kexin Zhang et al. \cite{zhang2024self} classify time series self-supervised learning (SSL) methods into three main categories: generative-based methods, contrastive-based methods, and adversarial-based methods. For contrastive-based methods, existing works \cite{chen2020simple, eldele2021time, eldele2023self, yue2022ts2vec, liu2024timesurl} focus on how positive and negative samples are generated based on data characteristics as Appendix~\ref{sec:appendix_a_1}.Most methods employ data augmentation techniques to generate different views of an input sample, learning representations by maximizing the similarity between views of the same sample while minimizing the similarity between views of different samples. However, the aforementioned methods rely on manually constructing positive and negative sample pairs. In contrast, adaptive contrastive learning has emerged as a promising direction in contrastive learning, exemplified by methods like LNT\cite{schneider2022detecting}, which introduces Dynamic Deterministic Contrastive Loss to enable models to adaptively learn diverse positive and negative sample representations. This approach is particularly beneficial for capturing the multi-scale nonlinear dynamics and individual variability inherent in medical time series data.
\section{Preliminaries and Problem Formulation}
\begin{figure*}[h]
\centering
\includegraphics[width=1\textwidth]{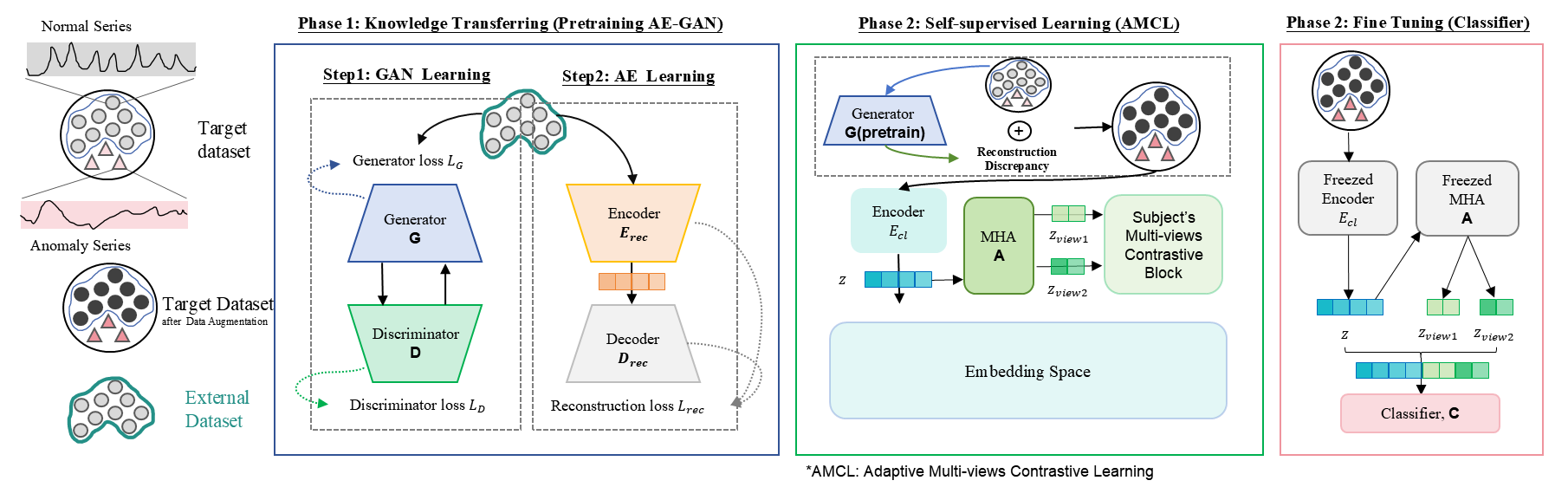}
\caption{Training Procedure of LMCRD}
\label{fig:train_proc}
\end{figure*}
Let the dataset of medical time series be denoted as \( \mathcal{D} \), where each series of data originates from a corresponding subject \( s \) and trial \( r \). Each subject \( s \) participates in multiple trials \( r \). Due to the lengthy nature of medical sequence data, they are segmented into several shorter sequence fragments. The segmented sequence samples \( \mathbf{\bar x}_{i} \in \mathbb{R}^{T \times F} \) are input to the model, with \( T \) denoting the sequence duration and \( F \) the feature dimension. To facilitate a deeper understanding, Appendix~\ref{sec:appendix_a_1} provides a detailed explanation of the hierarchical structuring of medical time series, illustrated using EEG signals from Alzheimer's disease as a case study. The external dataset \( \mathcal{D}' \) is collected for the same medical task as the target dataset \( \mathcal{D} \), but they originate from different centers. Nonetheless, they share similar feature characteristics, which ensures the applicability of \( \mathcal{D}' \) in the knowledge transfer to our target domain.

As shown in Figure \ref{fig:train_proc}, our training process consists of three stages. First, we pretrain an AE-GAN model using data from normal patients in \( \mathcal{D}' \), whose samples are expressed as \( \mathbf{x}'_{i} \). Next, we use the pretrained AE-GAN model to perform inference on all samples in \( \mathcal{D} \), whose samples are expressed as \( \mathbf{x}_{i} \), obtaining reconstruction error as additional feature. The feature is then merged with the raw data, resulting in a feature-augmented dataset with the number of features increased from \( F \) to \( F+1 \). This augmented dataset proceeds to the second stage of self-supervised learning. During this stage, through contrastive learning, we obtain an encoder \( E_{LMCRD} \) with a multi-head attention block and its embedding space is shown in Figure \ref{fig:embedding_space}. This encoder generates representations \( \mathbf{h}_{i} \in \mathbb{R}^{T \times C} \) with \( C \) dimensions, and view-level representations \( \mathbf{g}_{i} \in \mathbb{R}^{T \times V \times d} \) with \( V \) views and \( d \) dimensions for each view. These representations are then used in the third stage of supervised learning to fine-tune specific scenarios.
\section{Method}
To overcome the limitations in medical time-series analysis, such as data scarcity, heterogeneity, and class imbalance, we introduce a two-stage framework that combines anomaly detection with a learnable multi-view contrastive learning strategy.The first stage leverages an AE-GAN model to perform knowledge transfer by capturing the normal data distribution from external datasets and generating enriched feature representations, which mitigate heterogeneity and data limitations in the target dataset. Building on these augmented features, the second stage employs a Learnable Multi-Views Contrastive Framework (LMCF) to extract diverse and robust temporal representations through multi-level contrastive learning. This includes subject-wise, trial-wise, epoch-wise, and temporal-wise views, complemented by inter-view and intra-view contrastive mechanisms. The overall workflow of the framework is depicted in Appendix ~\ref{fig:overview} and detailed in the subsequent sections.
\subsection{AE-GAN}
To address the challenges of data scarcity and data heterogeneity in medical time series (\ref{sec:appendix_a_2}), we propose a knowledge transfer framework based on anomaly detection. This framework leverages knowledge from external datasets to enhance the feature representation of the target dataset, thereby improving the model's classification performance. Specifically, we introduce a hybrid model, AE-GAN \cite{larsen2016autoencoding, xu2022generative}, which combines a Generative Adversarial Network (GAN) and an Autoencoder (AE). The AE-GAN model learns the distribution of normal samples from the external dataset and generates high-quality feature representations for normal data. These representations are then used to perform feature augmentation on the target dataset, transferring prior knowledge about normal samples from the external dataset to the model. This approach effectively mitigates challenges such as limited data size, class imbalance, and multi-center distribution differences. In this way, our framework provides a more robust and generalizable model for medical diagnosis.
The AE-GAN model consists of two key components: a generator with an encoder-decoder architecture and a discriminator. The generator learns to reconstruct normal data by training on normal samples from the external dataset, while the discriminator distinguishes between real and generated samples. Through training on the external dataset, the model captures the underlying distribution of normal data and uses this knowledge to enhance the feature representation of the target dataset. This process not only enriches the feature space but also effectively addresses the issues of data scarcity and heterogeneity.
In the following sections, we will use the external dataset as an example to detail the structure of the AE-GAN model, its training process, and how it achieves knowledge transfer (or feature augmentation) for the target dataset.
\subsubsection{Generator}

The generator $G_{gen}$ is composed of an encoder $E_{gen}$ and a decoder $D_{gen}$, which work together to learn the distribution of healthy samples $\mathbf{x}'_i$ from the external dataset $D'$. In details, The encoder maps the healthy samples $\mathbf{x}'_i$ to a latent space representation and the decoder reconstructs the data from the representation to the generated data $\hat{\mathbf{x}}'_i$:

\begin{equation}
  \hat{\mathbf{x}}'_i = G_{gen}(\mathbf{x}'_i) = D_{gen}(E_{gen}(\mathbf{x}'_i))
\end{equation}

The generator is trained to minimize the reconstruction loss, ensuring that the generated data $\hat{\mathbf{x}}'_i$ closely approximates the original healthy samples $\mathbf{x}'_i$. This enables the generator to capture the underlying distribution of healthy data, which is critical for subsequent feature augmentation and knowledge transfer.

\subsubsection{Discriminator}

The discriminator $D_{dis}$ is responsible for distinguishing between real healthy samples $\mathbf{x}'_i$ from the external dataset $D'$ and generated data $\hat{\mathbf{x}}'_i$. The output of $D_{dis}$ represents the probability that the sample is from healthy subject. The details about the loss function can be seen in \ref{sec:appendix_d}.
\subsubsection{Inference on Internal Data}
After pretraining, we use the AE-GAN model to perform inference on all samples $\mathbf{\bar x}_i$ on our target dataset. The specific steps are as follows:

i. \textbf{Input Internal Data $\mathbf{\bar x}_i$}: The samples $\mathbf{\bar x}_i$ from the target dataset are fed into the pretrained generator $G_{gen}$.

ii. \textbf{Compute Reconstruction Error $\mathcal{E}$}: The generator $G$ reconstructs the samples $\mathbf{\bar x}_i$, and the reconstruction error $\mathcal{E}$ is computed as:

\vspace{-1em}
\begin{equation}
\mathcal{E} = \text{MSE}(G_{gen}(\mathbf{\bar x}_i), \mathbf{\bar x}_i)
\end{equation}
%\vspace{-2em}

   % \[
   % \epsilon = \text{MSE}(\text{Generator}(x_i), x_i)
   % \]
   Since the generator is pretrained on healthy samples $\mathbf{x}'_i$ from $D'$, the reconstruction error $\mathcal{E}$ is smaller for healthy samples $\mathbf{x}_i$ in the target dataset and larger for abnormal samples $\mathbf{x}_i$. Thus, the magnitude of $\mathcal{E}$ indirectly reflects the abnormality of the sample $\mathbf{x}_i$, i.e., the probability of disease.
   
iii. \textbf{Feature Augmentation}: The reconstruction error $\mathcal{E}$ is concatenated with the original features $x_i$ of the target dataset to form an augmented feature set:
\begin{equation}
\mathbf{x}_i = \text{concat}(\mathbf{\bar x}_i,\mathcal{E})
\end{equation}
   The augmented dataset $ {\mathbf{x}_i} $ is used for subsequent model development.
\subsection{Learnable Multi-views Contrastive Framework}
In the subsequent phase of our research, we harness the power of contrastive learning, a self-supervised learning technique, to generate series representations. The encoder $E_{LMCRD}$ of our approach is designed with two components.

\begin{figure}[htbp]
\centering
\vspace{5pt}  % 减少下间距
\includegraphics[width=0.5\columnwidth]{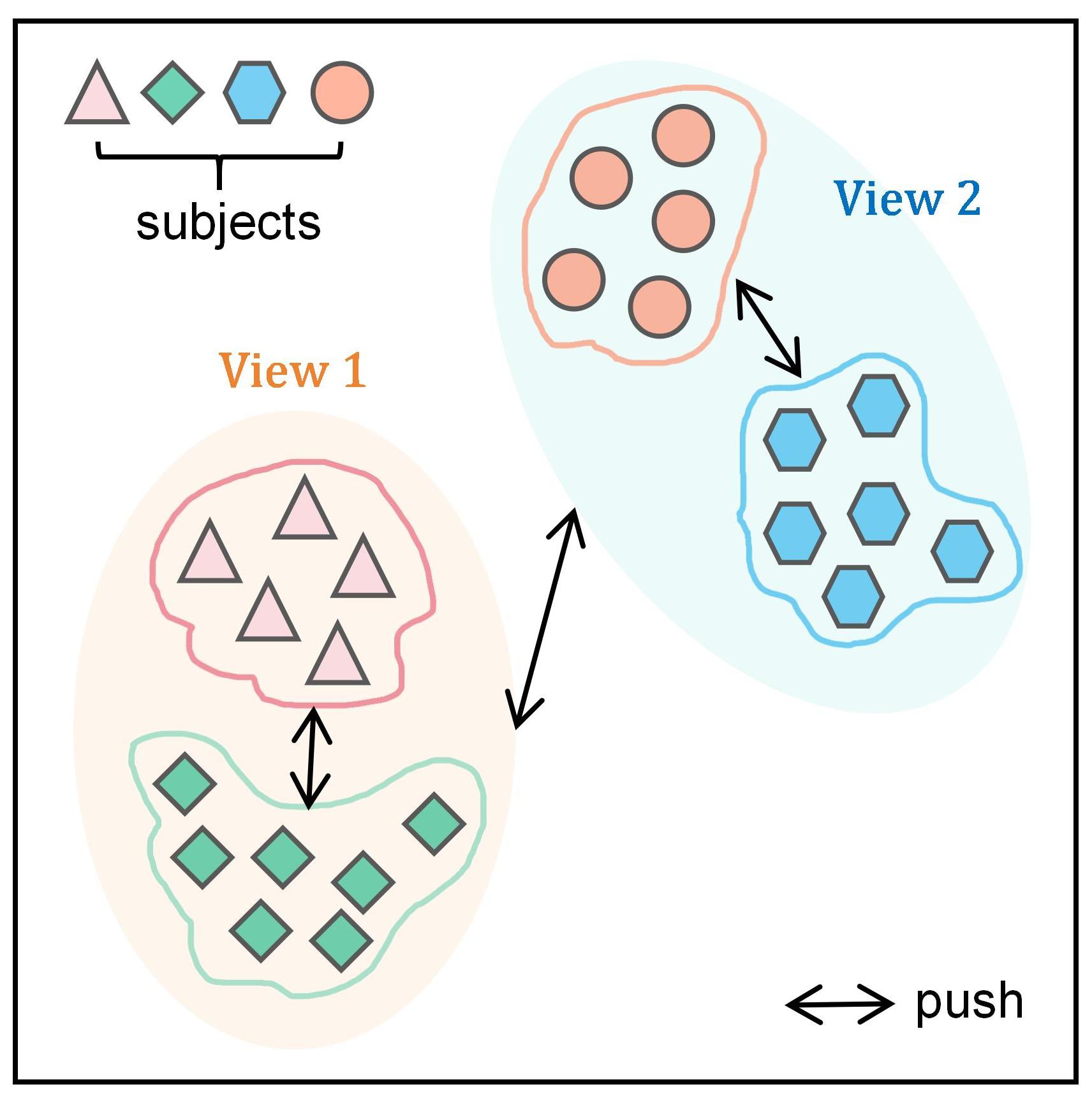}
% \vspace{-5pt}  % 减少下间距
\vspace{-10pt}  % 减少下间距
\caption{Mechanism of Multi-View Contrastive Learning}
\label{fig:view_cl}
% \vspace{-10pt}  % 减少下间距
\end{figure}

The first component is a dilated convolutional network $h(\mathbf{x})$, which broadens the receptive field of the sequence. This network is crucial for capturing a broader range of information and enhancing the generalizability of the temporal representations. On the other hand, inspired from the local transformations to craft diverse data views in the latent space for contrastive learning \cite{schneider2022detecting}, we recognized that MHA can significantly reduce the manual cost of designing positive and negative sample pairs from different perspectives of data. Consequently, we incorporated a second component into our encoder, which is our innovative learnable multi-views network $g(\mathbf{x})$. It introduces MHA to contrastive learning for extracting varying view representations adaptively. By leveraging both inter-wise and intra-wise contrastive losses, we ensure that the representations not only encompass diverse views but also exhibit distinct separability among different subjects within each view as Figure \ref{fig:view_cl} shown.

Furthermore, to capture the intrinsic hierarchical information embedded in medical data, we adopt the concept from COMET \cite{wang2024contrast}. Our contrastive learning considers subject, trial, epoch, and timestamp, and optimizes the model using multiple InfoNCE losses \cite{oord2018representation}. Ultimately, we obtain representations that are beneficial for downstream medical time-series tasks.
\vspace{-5pt}  % 减少下间距

\vspace{-10pt}  % 减少下间距
\textbf{Subject-wise Contrastive Loss} $L_{S}$: 

Subject may exhibit unique physiological characteristics and pathological variations. Consequently, the subject-wise contrastive learning is designed to enhance the discriminative representation of subject-specific features by increasing the similarity among samples from the same subject and the separability among samples from different subjects.Specifically, let \( \mathbf{x}_{i,s} \) denote the \( i \)-th anchor sample in the given batch, which is from subject \( s \). The positive sample set \( S_{i}^{+} \) for \( \mathbf{x}_{i,s} \) consists of all other samples \( \mathbf{x}_{j, s^{+}} \) with the same subject \( s^{+} \). Correspondingly, the negative sample set \( S_{i}^{-} \) includes all samples \( \mathbf{x}_{j, s^{-}} \) from different subjects \( s^{-} \). Based on this delineation of positive and negative samples, we define the subject-level contrastive loss function as follows:

\textbf{Subject-wise Contrastive Loss} $L_{S}$: 

Subject may exhibit unique physiological characteristics and pathological variations. Consequently, the subject-wise contrastive learning is designed to enhance the discriminative representation of subject-specific features by increasing the similarity among samples from the same subject and the separability among samples from different subjects.Specifically, let \( \mathbf{x}_{i,s} \) denote the \( i \)-th anchor sample in the given batch, which is from subject \( s \). The positive sample set \( S_{i}^{+} \) for \( \mathbf{x}_{i,s} \) consists of all other samples \( \mathbf{x}_{j, s^{+}} \) with the same subject \( s^{+} \). Correspondingly, the negative sample set \( S_{i}^{-} \) includes all samples \( \mathbf{x}_{j, s^{-}} \) from different subjects \( s^{-} \). Based on this delineation of positive and negative samples, we define the subject-level contrastive loss function as follows:

\textbf{Trial-wise Contrastive Loss} $L_{R}$:

Trials may demonstrate distinctive experimental conditions and variations in outcomes. To address this, trial-wise contrastive learning enhances trial-specific feature representation by promoting similarity within trials and separability between trials.

Concretely, assume \( \mathbf{x}_{i,r} \) represent the \( i \)-th anchor sample in the batch, sourced from trial \( r \). The set of positive samples \( R_{i}^{+} \) associated with \( \mathbf{x}_{i,r} \) encompasses all other samples \( \mathbf{x}_{j, r^{+}} \) that are from the same trial \( r^{+} \). Conversely, the set of negative samples \( R_{i}^{-} \) consists of all samples \( \mathbf{x}_{j, r^{-}} \) originating from different trials \( r^{-} \). Therefore, we establish the trial-wise contrastive loss function as below:
\vspace{1em}
\begin{equation}
\small
L_{R} = \frac{-1}{M} \sum_{i=1}^{M} \sum_{\mathbf{x}_{j,r^{+}} \in R_{i}^{+}} \log \frac{ \exp\left(s(\mathbf{h}_{i,r}, \mathbf{h}_{j,r^{+}}) / \tau\right)}{\sum_{\mathbf{x}_{j,r^{-}} \in R_{i}^{-}} \exp\left({s(\mathbf{h}_{i,r}, \mathbf{h}_{j,r^{-}}) / \tau}\right)} 
\end{equation}

where, \( \mathbf{h}_{i,r} \) denotes the representation obtained by applying the network \( h(\mathbf{x}) \) to the sample \( \mathbf{x}_{i,r} \), formally expressed as \( \mathbf{h}_{i,r} = h(\mathbf{x}_{i,r}) \). Consistent with this notation, \( \mathbf{h}_{j,r^{+}} \) and \( \mathbf{h}_{j,r^{-}} \) represent the feature representations resulting from the application of \( h(\mathbf{x}) \) to the corresponding positive and negative samples \( \mathbf{x}_{j,r^{+}} \) and \( \mathbf{x}_{j,r^{-}} \).

\textbf{Epoch-wise Contrastive Loss} $L_{E}$:

For epoch-wise contrastive learning, our hypothesis is that a sample which undergoes minor modifications, like temporal masking, should maintain a similar pattern compared to other sample. Hence, for each sample \( \mathbf{x}_i \) in batch, we employ a data augmentation strategy that is random continuous masking. This strategy involves applying two independent augmentations to \( \mathbf{x}_i \), resulting in augmented samples \( \tilde{\mathbf{x}}_{i}^{1} \) and \( \tilde{\mathbf{x}}_{i}^{2} \). Since these augmented samples originate from the same anchor sample \( \mathbf{x}_{i} \), they are encouraged to have high similarity between their feature representations and are considered as positive pair $(\tilde{\mathbf{x}}_{i}^{1}, \tilde{\mathbf{x}}_{i}^{2})$. In contrast, negative sample pairs are formed from augmented samples derived from different anchor samples, such as $\left ( \tilde{\mathbf{x}}_{i}^{1}, \tilde{\mathbf{x}}_{j}^{1} \right )$ and $\left ( \tilde{\mathbf{x}}_{i}^{1}, \tilde{\mathbf{x}}_{j}^{2} \right )$. Finally, the epoch-level loss function is formulated as:
\begin{equation}
% \tiny
\small
L_{E}=\frac{-1}{M}\sum_{i=1}^{M}\log\frac{\exp\left(\tilde{\mathbf{h}}_{i}^{1}\cdot \tilde{\mathbf{h}}_{i}^{2}\right)}{\sum\limits_{j=1,j\ne i}^{M}\left ( \exp\left ( \tilde{\mathbf{h}}_{i}^{1}\cdot \tilde{\mathbf{h}}_{j}^{1} \right )  +\exp\left (\tilde{\mathbf{h}}_{i}^{1}\cdot \tilde{\mathbf{h}}_{j}^{2}\right ) \right ) } 
\end{equation}

where $\tilde{\mathbf{h}}_{i}^{1}=h(\tilde{\mathbf{x}}_{i}^{1})$ and $\tilde{\mathbf{h}}_{i}^{2}=h(\tilde{\mathbf{x}}_{i}^{2})$.

\textbf{Temporal-wise Contrastive Loss} $L_{T}$:

The objective of this loss function is to ensure that two augmented samples originating from the same sample exhibit similar representations when observed at the same timestamp, while their representations differ when observed at other timestamps. Temporal-wise contrastive learning leverages the temporal consistency within the data. In this context, let us consider the anchor sample \(\mathbf{x}_{i,t}\) of the \(i\)-th data point at time \(t\) within a batch. By applying temporal masking, we generate two augmented samples as \(\tilde{\mathbf{x}}_{i,t}^{1}\) and \(\tilde{\mathbf{x}}_{i,t}^{2}\). The pair \((\tilde{\mathbf{x}}_{i,t}^{1}, \tilde{\mathbf{x}}_{i,t}^{2})\) is classified as a positive sample pair because they are derived from the same anchor sample at timestamp \(t\). Conversely, negative sample pairs are defined as \((\tilde{\mathbf{x}}_{i,t}^{1}, \tilde{\mathbf{x}}_{i,t^{-}}^{1})\) and \((\tilde{\mathbf{x}}_{i,t}^{1}, \tilde{\mathbf{x}}_{i,t^{-}}^{2})\), where \(t \neq t^{-}\). These negative pairs are constructed from augmented samples originating from different temporal points, thereby encouraging the model to learn distinct representations for different time observations. The Temporal-wise contrastive loss is shown as:
%\vspace{-1em}
% \begin{equation}
% \small
% L_{T}=\frac{-1}{T\cdot M}\sum_{t=1}^{T}\sum_{i=1}^{M}\log\frac{\exp\left(\tilde{\mathbf{h}}_{i,t}^{1}\cdot \tilde{\mathbf{h}}_{i,t}^{2}\right)}{\sum\limits_{t=1,t\ne t^{-}}^{T}\left ( \exp\left ( \tilde{\mathbf{h}}_{i,t}^{1}\cdot \tilde{\mathbf{h}}_{i,t^{-}}^{1} \right )  +\exp\left (\tilde{\mathbf{h}}_{i,t}^{1}\cdot \tilde{\mathbf{h}}_{i,t^{-}}^{2}\right ) \right ) } 
% \end{equation}
%\begin{equation}
\small
\begin{align}
L_{T} &= \frac{-1}{T\cdot M}\sum_{t=1}^{T }\sum_{i=1}^{M}\log Q_{i,t} \\
Q_{i,t} &= 
\frac{\exp\left(\tilde{\mathbf{h}}_{i,t}^{1}\cdot \tilde{\mathbf{h}}_{i,t}^{2}\right)}
{\sum\limits_{t=1,t\ne t^{-}}^{T}
\left( 
\exp\left ( \tilde{\mathbf{h}}_{i,t}^{1}\cdot \tilde{\mathbf{h}}_{i,t^{-}}^{1} \right )  
+\exp\left (\tilde{\mathbf{h}}_{i,t}^{1}\cdot \tilde{\mathbf{h}}_{i,t^{-}}^{2}\right ) 
\right)}
\end{align}
%\end{equation}

where $\tilde{\mathbf{h}}_{i,t}^{1}=h(\tilde{\mathbf{x}}_{i,t}^{1})$. Similarly, analogous notations with $h$ are derived from the network $h(\mathbf{x})$.

% view间
\textbf{Inter-view Contrastive Loss}: $L_{IRV}$: The inter-view contrastive loss we propose aims to encourage the model to capture informative features across different augmented views adaptively. Essentially, inter-view loss drives the representations from different views to be distinguishable from each other, thereby promoting the diversity of views.

Let us consider a batch of samples as $\mathbf{x}_i$, where $i = 1, \ldots, M$. For each sample $\mathbf{x}_i$, we apply data augmentation to generate two augmented samples, $\tilde{\mathbf{x}}_{i}^{1}$ and $\tilde{\mathbf{x}}_{i}^{2}$. These augmented samples are subsequently fed into the MHA module, denoted as $g(\mathbf{x})$. This module generates representations $\tilde{\mathbf{g}}_{i,v}^{k}$ for each augmented branch $k \in \left \{ 1,2 \right \}$ and different views $v$, where $g(\mathbf{x}_{i}^{k}) = \tilde{\mathbf{g}}_{(i,v)}^{k}$. The view representations from the given batch are then aggregated into $\tilde{\mathbf{g}}_{v}^{k} = \left \{ \tilde{\mathbf{g}}_{1,v}^{k}, \tilde{\mathbf{g}}_{2, v}^{k}, \ldots, \tilde{\mathbf{g}}_{M, v}^{k}\right \} $.

Assume we have two views for each branch that is $V=2$, To facilitate inter-view contrastive learning, we define positive sample pairs as those from the same view, such as $(\tilde{\mathbf{g}}_{1}^{1}, \tilde{\mathbf{g}}_{1}^{2})$ and $(\tilde{\mathbf{g}}_{2}^{1}, \tilde{\mathbf{g}}_{2}^{2})$. Conversely, negative sample pairs are those from different views, such as $(\tilde{\mathbf{g}}_{1}^{1}, \tilde{\mathbf{g}}_{2}^{1})$ and $(\tilde{\mathbf{g}}_{2}^{1}, \mathbf{g}_{1}^{2})$. Hence, the inter-view contrastive loss is formulated as:

\vspace{-1em}
\begin{equation}
L_{IRV} = \frac{-1}{V}\sum_{i=1}^{V}  \log \frac{\exp(\text{s}(\tilde{\mathbf{g}}_{i}^{1}, \tilde{\mathbf{g}}_{i}^{2}) )}{\sum_{j=1,j\ne i}^{V} \exp(\text{s}(\tilde{\mathbf{g}}_{i}^{1}, \tilde{\mathbf{g}}_{j}^{2}) )} 
\end{equation}
\vspace{-1em}

\textbf{Intra-View Contrastive Loss} $L_{IAV}$: In addition to inter-view loss, we also introduce the intra-view loss to further refine the representations within the same view. This loss is designed to enhance the discriminative power of representations by contrasting samples from the same view but different subjects, thus encouraging the model to capture subject-specific view features more effectively.

Similarly, assume we have subject-specific samples $\mathbf{x}_{i,s}$, where $i$ ranges from $1$ to $M$. For each of them, we apply data augmentation, resulting in augmented samples $\tilde{\mathbf{x}}_{i}^{1}$ and $\tilde{\mathbf{x}}_{i}^{2}$. These augmented samples are then processed through a MHA network $g(\mathbf{x})$, which produces subject-specific view representations $\tilde{\mathbf{g}}_{i,v}^{k}$ for each augmented branch $k \in \left \{ 1,2 \right \} $ and various views $v$, i.e., $\tilde{\mathbf{g}}_{i,v,s}^{k} = g(\mathbf{x}_{i,s}^{k})$.

Suppose we have $V$ views and $S$ subjects. To enable intra-view contrastive learning, we identify positive sample pairs as those from the same view and subject. Therefore, we have positive pairs that are $\left (  \tilde{\mathbf{g}}_{i,v,s}^{1}, \tilde{\mathbf{g}}_{i,v^{+},s^{+}}^{2}\right ) $. On the other hand, The negative sample pairs are defined as the samples that differ in subject at the given view, which are $\left (  \tilde{\mathbf{g}}_{i,v,s}^{1}, \tilde{\mathbf{g}}_{i,v^{+},s^{-}}^{2}\right ) $
\vspace{-1em}

\begin{equation}
\small
\begin{aligned}
L_{IAV} &= \frac{-1}{V\cdot M}\sum_{v=1}^{V}\sum_{i=1}^{M}
\sum_{\mathbf{x}_{j,v^{+},s^{+}} \in S_{i}^{+}} 
\log K_{i,v,s} \\
K_{i,v,s} &= \frac{\exp({s( \tilde{\mathbf{g}}_{i,v,s}^{1} \cdot \tilde{\mathbf{g}}_{j,v^{+},s^{+}}^{2})/\tau})}
{\sum\limits_{\mathbf{x}_{j,v^{a+},s^{-}} \in S_{i}^{-}}  
\exp({s(\tilde{\mathbf{g}}_{i,v,s}^{1} \cdot \tilde{\mathbf{g}}_{j,v^{+},s^{-}}^{2})/\tau })}
\end{aligned}
\end{equation}
where $S_{i}^{+}$ includes the augmented sample set of $\tilde{\mathbf{x}}_{j,v^{+},s^{+}}^{2}$ that belong to the same subject as the given augmented sample $\tilde{\mathbf{x}}_{i,v,s}^{2}$ under the same view $v$. Conversely, $S_{i}^{-}$ consists of the augmented samples that come from the same view but different subjects.

Thus, our view loss comprises both inter-view and intra-view contrastive losses:

\vspace{-1.5em}
\begin{equation}
\mathcal{L}_{V} = L_{IRV} + L_{IAV}
\end{equation}
\vspace{-2em}

Finally, the overall contrastive loss function $L$ is defined as:

\vspace{-1.5em} % 增加负间距
\begin{equation}
\mathcal{L} = \lambda_{S} \cdot L_{S} + \lambda_{R} \cdot L_{R} + \lambda_{E} \cdot L_{E} + \lambda_{T} \cdot L_{T} + \lambda_{V} \cdot L_{V}
\end{equation}
\vspace{-1.5em} % 增加负间距

where $\lambda_{S}:\lambda_{R}:\lambda_{E}:\lambda_{T}:\lambda_{V}=1:1:1:1:2$ in our practice.

\subsection{Fine Tuninig}
In order to use the Encoder $E_{LMCRD}$ of LMCRD for downstream tasks, we utilize fine-tuning as follows:
%\vspace{-2em}
\begin{align}
\hat{y}_{i} = C(E_{LMCRD}\left (  \mathbf{x}_{i} \right ) ) = C(h(\mathbf{x}_{i}), g(\mathbf{x}_{i}))
\end{align}
\vspace{-2em}

where $\hat{y}$ is the classification result for downstream medical disease diagnosis, and the training optimization objective is binary cross-entropy.

\section{Experiments}

\begin{figure*}[t]
\centering
\includegraphics[width=\textwidth]{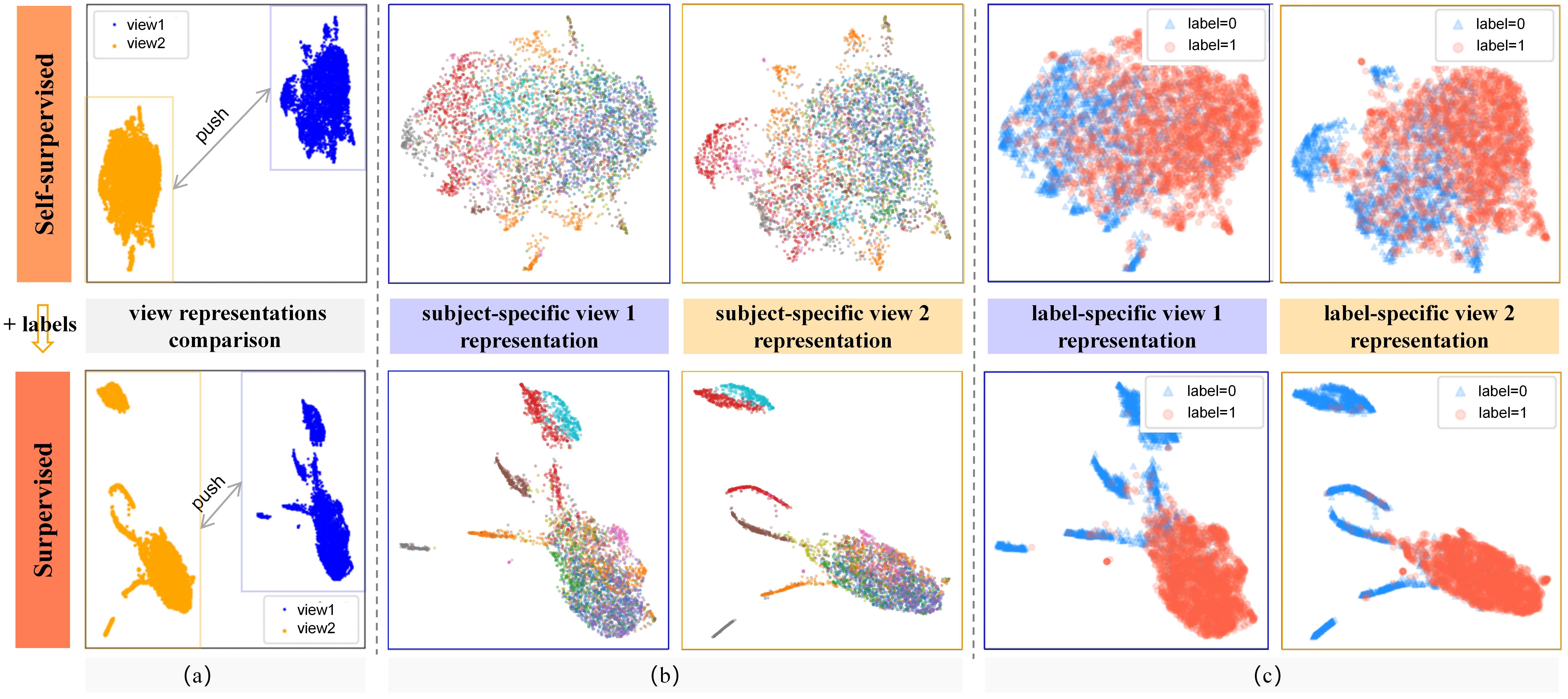}
\vspace{-20pt}
\caption{Visualization of Multi-View Representations on the AD Dataset. The figure compares contrastive view representations obtained through self-supervised contrastive learning and 100\% supervised fine-tuning.}
\vspace{-10pt}
\label{fig:view_representation}

\end{figure*}
\subsection{Datasets}
For Alzheimer’s disease, myocardial infarction, and Parkinson’s disease, we introduce three target datasets corresponding to each medical disease and three additional external datasets from other centers. By incorporating these external datasets, we aim to learn medical expertise from other centers to enhance the performance of downstream tasks on the target datasets. The detailed descriptions of these six datasets and the data preprocessing steps are presented in Appendix \ref{sec:appendix_c}. During the data splitting, we use the subject identifier as the criterion for division.
\subsection{Results}
\subsubsection{Results on Partial Fine-tuning}
To evaluate the representation quality of our proposed LMCRD quickly, we adopt the Partial Fine-Tuning (PFT). PFT refers to combining the Encoder with frozen parameters and a trainable logistic regression linear classifier $C$. In this setup, we utilizes 100\% of the labeled data for training the supervised classification. The PFT results on the AD dataset are shown in Table~\ref{tab:Partial Fine-tuning Results.}. Our proposed LMCRD model excels in all metrics, including Accuracy, Precision, Recall, F1 score, AUROC, and AUPRC, demonstrating the power of our representation learning.

\subsubsection{Results on Full Fine-tuning}
Unlike PFT, Full Fine-tuning (FFT) adapts all model parameters to the given target medical task, which means the Encoder and the classifier are trained together with supervision. FFT further enhances the representation in the downstream task.
The results of FFT on the target dataset are exemplified using the AD dataset in Table ~\ref{tab:performance_comparison}, with the remaining results provided in Appendix Table~\ref{tab:performance_comparison2}.

\begin{table*}[t]
\centering
\caption{Partial Fine-tuning Results.}
\label{tab:Partial Fine-tuning Results.}
\resizebox{0.7\textwidth}{!}{% 调整表格宽度以适应页面宽度
\begin{tabular}{l|c|c|c|c|c|c}
\hline
\textbf{Models} & \textbf{Accuracy} & \textbf{Precision} & \textbf{Recall} & \textbf{F1 score} & \textbf{AUROC} & \textbf{AUPRC} \\ \hline
\textbf{TS2vec} & $66.48\pm5.33$ & $67.72\pm5.09$ & $67.40\pm5.20$ & $66.32\pm5.46$ & $74.12\pm6.88$ & $72.96\pm7.21$ \\
\textbf{TF-C} & $77.03\pm2.80$ & $75.79\pm5.07$ & $64.27\pm5.03$ & $64.85\pm5.56$ & $80.71\pm4.03$ & $79.27\pm4.15$ \\
\textbf{Mixing-up} & $46.16\pm1.38$ & $52.62\pm4.90$ & $50.81\pm1.32$ & $37.37\pm1.98$ & $64.42\pm6.49$ & $62.85\pm6.07$ \\
\textbf{TS-TCC} & $59.71\pm8.63$ & $61.66\pm8.63$ & $60.33\pm3.26$ & $58.66\pm8.39$ & $67.53\pm10.04$ & $68.33\pm9.37$ \\
\textbf{SimCLR} & $57.16\pm2.05$ & $56.67\pm3.91$ & $53.57\pm2.21$ & $49.11\pm4.26$ & $56.67\pm3.91$ & $52.10\pm1.41$ \\
\textbf{CLOCS} & $66.99\pm2.84$ & $67.17\pm2.96$ & $67.33\pm2.99$ & $66.91\pm2.88$ & $73.71\pm3.62$ & $72.58\pm3.54$ \\
\textbf{COMET} & $76.09\pm4.21$ & $77.36\pm3.97$ & $74.68\pm4.62$ & $74.80\pm4.83$ & $81.30\pm4.97$ & $80.50\pm5.31$ \\ \hline
\textbf{LMCRD} & $\textbf{80.48}\pm\textbf{5.97}$ & $\textbf{80.34}\pm\textbf{6.00}$ & $\textbf{80.04}\pm\textbf{6.27}$ & $\textbf{80.12}\pm\textbf{6.22}$ & $\textbf{85.96}\pm\textbf{7.17}$ & $\textbf{85.40}\pm\textbf{7.52}$ \\
\hline
\end{tabular}
}
\end{table*}
\begin{table*}[t]
\centering
\caption{Full fine-tuning results of AD datasets. We use 100\% and 10\% of labeled data for fine-tuning. * presents our approach.}
\label{tab:performance_comparison}
\resizebox{0.8\textwidth}{!}{% 调整表格宽度以适应页面宽度
\begin{tabular}{c|clllllll}
\hline
\textbf{Datasets} & \textbf{Fraction} & \textbf{Models} & \textbf{Accuracy} & \textbf{Precision} & \textbf{Recall} & \textbf{F1 score} & \textbf{AUROC} & \textbf{AUPRC} \\ \hline
\multirow{18}{*}{\textbf{AD}} & \multirow{8}{*}{\textbf{100\%}} & \textbf{TS2vec} & $81.26 \pm 2.08$ & $81.21 \pm 2.14$ & $81.34 \pm 2.04$ & $81.12 \pm 2.06$ & $89.20 \pm 1.76$ & $88.94 \pm 1.85$ \\
 &  & \textbf{TF-C} & $75.31 \pm 8.27$ & $75.87 \pm 8.73$ & $74.83 \pm 8.98$ & $74.54 \pm 8.85$ & $79.45 \pm 10.23$ & $79.33 \pm 10.57$ \\ 
 &  & \textbf{Mixing-up} & $65.68 \pm 7.89$ & $72.61 \pm 4.21$ & $68.25 \pm 6.97$ & $63.98 \pm 9.92$ & $84.63 \pm 5.04$ & $83.46 \pm 5.48$ \\ 
 &  & \textbf{TS-TCC} & $73.55 \pm 10.00$ & $77.22 \pm 6.13$ & $73.83 \pm 9.65$ & $71.86 \pm 11.59$ & $86.17 \pm 5.11$ & $85.73 \pm 5.11$ \\ 
 &  & \textbf{SimCLR} & $54.77 \pm 1.97$ & $50.15 \pm 7.02$ & $50.58 \pm 1.92$ & $43.18 \pm 4.27$ & $50.15 \pm 7.02$ & $50.42 \pm 1.06$ \\ 
 &  & \textbf{CLOCS} & $78.37 \pm 6.00$ & $83.99 \pm 2.11$ & $76.14 \pm 7.03$ & $75.78 \pm 7.93$ & $91.17 \pm 2.51$ & $90.72 \pm 3.05$ \\ 
 &  & \textbf{COMET} & $84.50 \pm 4.46$ & $88.31 \pm 2.42$ & $82.95 \pm 5.39$ & $83.33 \pm 5.15$ & $94.44 \pm 2.37$ & $94.43 \pm 2.48$ \\ 
 &  & \textbf{LMCRD*} & $ \textbf{93.23}\pm \textbf{5.25}$ & $\textbf{94.01} \pm \textbf{3.98}$ & $\textbf{92.71} \pm \textbf{5.86}$ & $\textbf{92.97} \pm \textbf{5.65}$ & $\textbf{98.03} \pm \textbf{1.71}$ & $\textbf{98.03} \pm \textbf{1.75}$ \\ 
  \cline{2-9} 
 & \multirow{8}{*}{\textbf{10\%}} & \textbf{TS2vec} & $73.28 \pm 4.34$ & $74.14 \pm 4.33$ & $73.52 \pm 3.77$ & $73.00 \pm 4.18$ & $81.66 \pm 5.20$ & $81.58 \pm 5.11$ \\ 
 &  & \textbf{TF-C} & $75.66 \pm 11.21$ & $75.48 \pm 11.48$ & $75.58 \pm 11.59$ & $75.38 \pm 11.47$ & $81.38 \pm 14.19$ & $81.56 \pm 13.68$ \\
 &  & \textbf{Mixing-up} & $59.38 \pm 3.33$ & $64.85 \pm 4.38$ & $61.94 \pm 3.42$ & $58.17 \pm 3.41$ & $75.02 \pm 6.14$ & $73.44 \pm 5.82$ \\ 
 &  & \textbf{TS-TCC} & $77.83 \pm 6.90$ & $79.73 \pm 7.49$ & $76.18 \pm 7.21$ & $76.43 \pm 7.56$ & $84.12 \pm 7.32$ & $84.12 \pm 7.61$ \\ 
 &  & \textbf{SimCLR} & $56.09 \pm 2.25$ & $53.81 \pm 5.74$ & $51.73 \pm 2.39$ & $44.10 \pm 4.84$ & $53.81 \pm 5.74$ & $51.08 \pm 1.33$ \\  
 &  & \textbf{CLOCS} & $76.97 \pm 3.01$ & $81.70 \pm 3.21$ & $74.69 \pm 3.26$ & $74.75 \pm 3.61$ & $86.91 \pm 3.61$ & $86.70 \pm 3.64$ \\ 
 &  & \textbf{COMET} & $91.43 \pm 3.12$ & $92.52 \pm 2.36$ & $90.71 \pm 3.56$ & $91.14 \pm 3.31$ & $96.44 \pm 2.84$ & $96.48 \pm 2.82$ \\
&  & \textbf{LMCRD*} & $ \textbf{94.67}\pm \textbf{1.84}$ & $\textbf{94.93} \pm \textbf{1.76}$ & $\textbf{94.41} \pm \textbf{1.99}$ & $\textbf{94.58} \pm \textbf{1.89}$ & $\textbf{98.10} \pm \textbf{1.20}$ & $\textbf{98.19} \pm \textbf{1.15}$ \\ 
 \cline{2-9} \hline
\end{tabular}
}
\end{table*}

\begin{table*}[t]
\centering
\caption{Full fine-tuning results of AD datasets. We use 100\% and 10\% of labeled data for fine-tuning. * presents our approach.}
\label{tab:performance_comparison}
\resizebox{0.8\textwidth}{!}{% 调整表格宽度以适应页面宽度
\begin{tabular}{c|clllllll}
\hline
\textbf{Datasets} & \textbf{Fraction} & \textbf{Models} & \textbf{Accuracy} & \textbf{Precision} & \textbf{Recall} & \textbf{F1 score} & \textbf{AUROC} & \textbf{AUPRC} \\ \hline
\multirow{18}{*}{\textbf{AD}} & \multirow{8}{*}{\textbf{100\%}} & \textbf{TS2vec} & $81.26 \pm 2.08$ & $81.21 \pm 2.14$ & $81.34 \pm 2.04$ & $81.12 \pm 2.06$ & $89.20 \pm 1.76$ & $88.94 \pm 1.85$ \\
 &  & \textbf{TF-C} & $75.31 \pm 8.27$ & $75.87 \pm 8.73$ & $74.83 \pm 8.98$ & $74.54 \pm 8.85$ & $79.45 \pm 10.23$ & $79.33 \pm 10.57$ \\ 
 &  & \textbf{Mixing-up} & $65.68 \pm 7.89$ & $72.61 \pm 4.21$ & $68.25 \pm 6.97$ & $63.98 \pm 9.92$ & $84.63 \pm 5.04$ & $83.46 \pm 5.48$ \\ 
 &  & \textbf{TS-TCC} & $73.55 \pm 10.00$ & $77.22 \pm 6.13$ & $73.83 \pm 9.65$ & $71.86 \pm 11.59$ & $86.17 \pm 5.11$ & $85.73 \pm 5.11$ \\ 
 &  & \textbf{SimCLR} & $54.77 \pm 1.97$ & $50.15 \pm 7.02$ & $50.58 \pm 1.92$ & $43.18 \pm 4.27$ & $50.15 \pm 7.02$ & $50.42 \pm 1.06$ \\ 
 &  & \textbf{CLOCS} & $78.37 \pm 6.00$ & $83.99 \pm 2.11$ & $76.14 \pm 7.03$ & $75.78 \pm 7.93$ & $91.17 \pm 2.51$ & $90.72 \pm 3.05$ \\ 
 &  & \textbf{COMET} & $84.50 \pm 4.46$ & $88.31 \pm 2.42$ & $82.95 \pm 5.39$ & $83.33 \pm 5.15$ & $94.44 \pm 2.37$ & $94.43 \pm 2.48$ \\ 
 &  & \textbf{LMCRD*} & $ \textbf{93.23}\pm \textbf{5.25}$ & $\textbf{94.01} \pm \textbf{3.98}$ & $\textbf{92.71} \pm \textbf{5.86}$ & $\textbf{92.97} \pm \textbf{5.65}$ & $\textbf{98.03} \pm \textbf{1.71}$ & $\textbf{98.03} \pm \textbf{1.75}$ \\ 
  \cline{2-9} 
 & \multirow{8}{*}{\textbf{10\%}} & \textbf{TS2vec} & $73.28 \pm 4.34$ & $74.14 \pm 4.33$ & $73.52 \pm 3.77$ & $73.00 \pm 4.18$ & $81.66 \pm 5.20$ & $81.58 \pm 5.11$ \\ 
 &  & \textbf{TF-C} & $75.66 \pm 11.21$ & $75.48 \pm 11.48$ & $75.58 \pm 11.59$ & $75.38 \pm 11.47$ & $81.38 \pm 14.19$ & $81.56 \pm 13.68$ \\
 &  & \textbf{Mixing-up} & $59.38 \pm 3.33$ & $64.85 \pm 4.38$ & $61.94 \pm 3.42$ & $58.17 \pm 3.41$ & $75.02 \pm 6.14$ & $73.44 \pm 5.82$ \\ 
 &  & \textbf{TS-TCC} & $77.83 \pm 6.90$ & $79.73 \pm 7.49$ & $76.18 \pm 7.21$ & $76.43 \pm 7.56$ & $84.12 \pm 7.32$ & $84.12 \pm 7.61$ \\ 
 &  & \textbf{SimCLR} & $56.09 \pm 2.25$ & $53.81 \pm 5.74$ & $51.73 \pm 2.39$ & $44.10 \pm 4.84$ & $53.81 \pm 5.74$ & $51.08 \pm 1.33$ \\  
 &  & \textbf{CLOCS} & $76.97 \pm 3.01$ & $81.70 \pm 3.21$ & $74.69 \pm 3.26$ & $74.75 \pm 3.61$ & $86.91 \pm 3.61$ & $86.70 \pm 3.64$ \\ 
 &  & \textbf{COMET} & $91.43 \pm 3.12$ & $92.52 \pm 2.36$ & $90.71 \pm 3.56$ & $91.14 \pm 3.31$ & $96.44 \pm 2.84$ & $96.48 \pm 2.82$ \\
&  & \textbf{LMCRD*} & $ \textbf{94.67}\pm \textbf{1.84}$ & $\textbf{94.93} \pm \textbf{1.76}$ & $\textbf{94.41} \pm \textbf{1.99}$ & $\textbf{94.58} \pm \textbf{1.89}$ & $\textbf{98.10} \pm \textbf{1.20}$ & $\textbf{98.19} \pm \textbf{1.15}$ \\ 
 \cline{2-9} \hline
\end{tabular}
}
\end{table*}

\subsection{Visualization of Views}
To verify the learnable multi-view learning capability of LMCF, we employed the UMAP \cite{mcinnes2018umap} to reduce the dimensionality of view representations to two dimensions and visualize them as shown in Figure \ref{fig:view_representation}. The visualizations provide strong validation from the following two aspects:

\textbf{The Disparities between Diverse Views}: By comparing the view representations as Figure \ref{fig:view_representation} (a), it can be seen that the designed inter-view contrastive loss enables the representations of view 1 and view 2 under different samples to show obvious disparities, effectively distinguishing the representations of each view.

\textbf{The Intra-view Subject Discrepancies}: The intra-view contrastive loss we propose fully considers the subject identity, setting the positive samples as those samples with the same view and subject as the anchor sample, and the negative samples as those with the same view but different subjects with the anchor sample. From the visualization results of patient-specific view representations in Figure \ref{fig:view_representation} (b), it can be found that there are differences in the representations of some different subjects under each view. In the self-supervised part, although label is not introduced, when observing the representations of different views with label coloring in Figure \ref{fig:view_representation} (c), the representation differences between diseased subjects and normal subjects in the AD dataset can still be found. This indicates that the view representations have good generalization ability, which is beneficial for downstream disease diagnosis tasks.

In addition, Figure \ref{fig:view_representation} also shows that after adding 100\% labeled data for supervised fine-tuning, the clustering of view representations is further enhanced. Overall, through this visualization, it is sufficient to prove the effectiveness of our LMCF in the adaptive contrastive learning.

\section{Conclusion}
In medical time-series tasks, models often overfit due to the lack of labeled data. To address this issue, we introduce external normal subject data and use AE-GAN to obtain the reconstruction discrepancy of the input sequence. This discrepancy is recognized as prior knowledge for measuring the probability of illness and is incorporated into the self-supervised contrastive learning phase. In this stage, we propose a Learnable Multi-views Contrastive Framework. By introducing multi-head attention mechanisms and view contrast learning modules, this contrastive model can adaptively learn representations of different views based on medical data characteristics, reducing the need for manual organization of positive and negative samples. Moreover, the design of inter-view and intra-view losses ensures diversity among view representations while preserving the differences between subjects within each view. Through fine-tuning downstream tasks, our experiments show that the model generates informative representations and achieves state-of-the-art performance on the AD, PTB, and TDBrain medical datasets.

\bibliographystyle{abbrv}  % 或者 plain, abbrv, unsrt, etc.
\bibliography{references}  
\newpage
\appendix
\section{Preliminary Knowledge on Medical Time Series}
\subsection{Hierarchical Structuring of Medical Time Series}
\label{sec:appendix_a_1}
To better understand the structure of medical time series data, this section provides a comprehensive explanation of its hierarchical organization, using EEG signals from Alzheimer’s disease as an example. The dataset is categorized into four hierarchical levels: Subject, Trial, Epoch, and Temporal, with each level representing a unique granularity of the data (Appendix figure~\ref{fig:sample_image}). Unlike conventional time series, which typically consist of sample and observation levels, medical time series incorporate two additional layers: patient and trial. These extra layers enhance the dataset's structure, enabling the development of methods tailored to the specific characteristics and analytical requirements of medical time series. As illustrated in Appendix Table~\ref{tab:models_levels}, many existing approaches utilize only a subset of these levels, underscoring the importance of leveraging the complete hierarchy for more effective analysis.

\textbf{Definition 1:Subject}

A \textbf{subject} \( \mathbf{p}_j \in \mathbb{R}^G \) in medical time series data represents the collection of multiple trials corresponding to a single individual. The subscript \( j \) denotes the subject ID. Trials associated with the same subject may differ due to variations in data collection timeframes, sensor placement, or patient conditions. A subject's data is generally represented as an aggregate of trials. Each trial within a subject \( \mathbf{p}_j \) is typically segmented into smaller samples to facilitate representation learning. To denote all the samples belonging to a specific subject \( \mathbf{p}_j \), we use the notation \( \mathbf{P}_j \).

\textbf{Definition 2: Trial}

A \textbf{trial} \( \mathbf{r}_k \) refers to a continuous set of observations collected over an extended time period (e.g., 30 minutes). This can also be referred to as a \textbf{record}. The subscript \( k \) denotes the trial ID. Trials are generally too lengthy (e.g., consisting of hundreds of thousands of observations) to be directly processed by deep learning models. Therefore, they are usually divided into smaller subsequences, referred to as samples or segments. The collection of all samples derived from a trial \( \mathbf{r}_k \) is denoted by \( \mathbf{R}_k \).

\textbf{Definition 3: Epoch}

An \textbf{epoch} \( e_m \) is a sequence of consecutive observations derived from a trial, typically spanning a shorter time period (e.g., several seconds). In EEG (electroencephalography), an epoch refers to a specific time window extracted from the continuous EEG signal. These time windows are used to isolate signal segments for further analysis, often aligned with specific events or experimental conditions. Each epoch is composed of multiple observations measured at regular intervals over a time range defined by \( M \) timestamps. To denote a specific epoch within a trial, we use \( e_m \), where \( m \) represents the epoch index. The set of observations forming an epoch is \[e_m = \{ e_{m,t} \mid t = 1, \cdots, M \}.\]

\textbf{Definition 4: Temporal}

A \textbf{temporal observation} \( \mathbf{x}_{n,t} \in \mathbb{R}^H \) refers to a single data point or vector recorded at a specific timestamp \( t \). The subscript \( n \) denotes the sample index, while \( t \) represents the timestamp. For univariate time series, the temporal observation is a single real value, whereas, for multivariate time series, it is a vector with \( H \) dimensions. Temporal observations may represent physiological signals, laboratory measurements, or other health-related indicators.

\subsection{Data Scarcity and Heterogeneity for Medical Time Series}
\label{sec:appendix_a_2}

\textbf{Data Scarcity}
Medical data, especially high-quality neurodegenerative disease data, is often expensive, with high costs associated with data collection and annotation. Additionally, individual datasets typically have limited sample sizes and suffer from class imbalance issues (e.g., an unequal ratio of healthy individuals to patients). These problems lead to overfitting during model training, poor generalization, and suboptimal classification performance.

\textbf{Data Heterogeneity}
Medical data often exhibits a multi-center characteristic, meaning datasets from different institutions show significant distributional differences. This heterogeneity makes it difficult to directly transfer models trained on one dataset to others, severely hindering the development and application of robust deep learning models.

\section{Supplementary Figures and Tables}
\label{sec:appendix_b}
\renewcommand{\thetable}{B\arabic{table}}
\renewcommand{\thefigure}{B\arabic{figure}} 
\setcounter{table}{0}
\setcounter{figure}{0}
\begin{table}[htbp]
\centering
\caption{\textbf{Different Models Utilize Different Levels}}
\label{tab:models_levels} % 标签放在标题后
\small  % 设置表格字体为小号
\begin{tabular}{lccccc}  % 使用 l 表示列左对齐
\hline
\textbf{Models} & \textbf{Subject} & \textbf{Trial} & \textbf{Epoch} & \textbf{Temporal} & \textbf{View} \\ \hline
\textbf{SimCLR} & & & \checkmark & & \\
\textbf{TF-C} & & & \checkmark & & \\
\textbf{Mixing-up} & & & \checkmark & & \\
\textbf{TNC} & & \checkmark & & & \\
\textbf{TS2vec} & & & \checkmark & \checkmark & \\
\textbf{TS-TCC} & & & \checkmark & \checkmark & \\
\textbf{CLOCS} & \checkmark & & \checkmark & & \\
\textbf{COMET} & \checkmark & \checkmark & \checkmark & \checkmark & \\ \hline
\textbf{LMCRD} & \checkmark & \checkmark & \checkmark & \checkmark & \textbf{\checkmark} \\ \hline
\end{tabular}
\end{table}

\begin{table*}[t]
\centering
\caption{Full fine-tuning results of two medical datasets. We use 100\% and 10\% of labeled data for fine-tuning. * presents our approach.}
\label{tab:performance_comparison2}
\resizebox{0.8\textwidth}{!}{% 调整表格宽度以适应页面宽度
\begin{tabular}{c|clllllll}
\hline
\textbf{Datasets} & \textbf{Fraction} & \textbf{Models} & \textbf{Accuracy} & \textbf{Precision} & \textbf{Recall} & \textbf{F1 score} & \textbf{AUROC} & \textbf{AUPRC} \\ \hline
\multirow{18}{*}{\textbf{TDBrain}} & \multirow{8}{*}{\textbf{100\%}} & \textbf{TS2vec} & $80.21 \pm 1.69$ & $81.38 \pm 1.97$ & $80.21 \pm 1.69$ & $80.07 \pm 1.69$ & $89.57 \pm 2.31$ & $89.60 \pm 2.37$ \\
 &  & \textbf{TF-C} & $66.62 \pm 1.76$ & $67.15 \pm 1.64$ & $66.62 \pm 1.76$ & $66.35 \pm 1.91$ & $65.43 \pm 6.13$ & $66.18 \pm 4.90$ \\ 
 &  & \textbf{Mixing-up} & $81.47 \pm 1.07$ & $82.11 \pm 1.12$ & $81.47 \pm 1.07$ & $81.38 \pm 1.08$ & $90.48 \pm 0.89$ & $90.51 \pm 0.04$ \\ 
 &  & \textbf{TS-TCC} & $77.42 \pm 2.86$ & $77.68 \pm 2.93$ & $77.42 \pm 2.86$ & $77.37 \pm 2.86$ & $87.37 \pm 3.06$ & $87.61 \pm 3.22$ \\ 
 &  & \textbf{SimCLR} & $58.43 \pm 1.77$ & $59.48 \pm 1.95$ & $58.43 \pm 1.77$ & $57.30 \pm 2.07$ & $59.48 \pm 1.95$ & $55.05 \pm 1.18$ \\ 
 &  & \textbf{CLOCS} & $78.16 \pm 1.13$ & $79.49 \pm 1.61$ & $78.16 \pm 1.13$ & $77.91 \pm 1.12$ & $88.49 \pm 1.95$ & $88.83 \pm 1.94$ \\ 
 &  & \textbf{COMET} & $85.47 \pm 1.16$ & $85.68 \pm 1.20$ & $85.47 \pm 1.16$ & $85.45 \pm 1.16$ & $93.73 \pm 1.02$ & $93.96 \pm 0.99$ \\ 
 % &  & \textbf{LMCF*} & $ \textbf{93.26}\pm \textbf{2.33}$ & $\textbf{92.84} \pm \textbf{1.76}$ & $\textbf{93.44} \pm \textbf{1.62}$ & $\textbf{93.07} \pm \textbf{1.92}$ & $\textbf{96.74} \pm \textbf{1.98}$ & $\textbf{97.60} \pm \textbf{1.82}$ \\
 %  &  & \textbf{COMET+AE-GAN*} & $ \textbf{90.61}\pm \textbf{1.78}$ & $\textbf{90.56} \pm \textbf{1.49}$ & $\textbf{90.16} \pm \textbf{1.94}$ & $\textbf{91.55} \pm \textbf{2.38}$ & $\textbf{95.69} \pm \textbf{2.07}$ & $\textbf{96.07} \pm \textbf{1.85}$ \\ 
 &  & \textbf{LMCRD*} & $ \textbf{95.60}\pm \textbf{1.20}$ & $\textbf{94.95} \pm \textbf{1.69}$ & $\textbf{95.73} \pm \textbf{1.74}$ & $\textbf{94.61} \pm \textbf{1.26}$ & $\textbf{97.63} \pm \textbf{1.57}$ & $\textbf{98.51} \pm \textbf{1.29}$ \\ \cline{2-9} 
 & \multirow{8}{*}{\textbf{10\%}} & \textbf{TS2vec} & $72.39 \pm 1.13$ & $74.49 \pm 1.73$ & $72.39 \pm 1.13$ & $71.80 \pm 1.05$ & $80.71 \pm 1.90$ & $80.06 \pm 1.87$ \\ 
 &  & \textbf{TF-C} & $59.14 \pm 3.04$ & $59.34 \pm 3.19$ & $59.14 \pm 3.04$ & $58.98 \pm 2.94$ & $59.56 \pm 4.10$ & $59.65 \pm 2.99$ \\
 &  & \textbf{Mixing-up} & $77.50 \pm 2.07$ & $80.09 \pm 1.92$ & $77.50 \pm 2.07$ & $76.99 \pm 2.28$ & $87.29 \pm 1.34$ & $87.13 \pm 1.37$ \\ 
 &  & \textbf{TS-TCC} & $71.23 \pm 1.57$ & $78.78 \pm 0.66$ & $71.23 \pm 1.57$ & $69.18 \pm 2.25$ & $80.56 \pm 1.98$ & $80.21 \pm 2.21$ \\ 
 &  & \textbf{SimCLR} & $59.79 \pm 2.09$ & $60.50 \pm 1.90$ & $59.79 \pm 2.09$ & $59.06 \pm 2.82$ & $60.50 \pm 1.90$ & $55.96 \pm 1.41$ \\  
 &  & \textbf{CLOCS} & $75.04 \pm 2.65$ & $75.97 \pm 2.86$ & $75.04 \pm 2.65$ & $74.83 \pm 2.66$ & $84.25 \pm 3.27$ & $84.37 \pm 3.52$ \\ 
 &  & \textbf{COMET} & $79.28 \pm 4.64$ & $79.83 \pm 4.83$ & $79.28 \pm 4.64$ & $79.19 \pm 4.62$ & $88.39 \pm 4.13$ & $88.38 \pm 3.96$ \\
 % &  & \textbf{LMCRD-LMC*} & $ \textbf{82.20}\pm \textbf{3.08}$ & $\textbf{81.53} \pm \textbf{3.91}$ & $\textbf{81.90} \pm \textbf{3.51}$ & $\textbf{82.09} \pm \textbf{3.60}$ & $\textbf{93.73} \pm \textbf{3.27}$ & $\textbf{94.81} \pm \textbf{3.49}$ \\
 %  &  & \textbf{LMCRD-RD*} & $ \textbf{81.27}\pm \textbf{4.32}$ & $\textbf{81.92} \pm \textbf{4.29}$ & $\textbf{82.30} \pm \textbf{4.33}$ & $\textbf{82.07} \pm \textbf{4.52}$ & $\textbf{90.22} \pm \textbf{4.02}$ & $\textbf{90.50} \pm \textbf{4.09}$ \\ 
 &  & \textbf{LMCRD*} & $ \textbf{83.66}\pm \textbf{3.29}$ & $\textbf{84.06} \pm \textbf{3.06}$ & $\textbf{83.29} \pm \textbf{2.96}$ & $\textbf{84.60} \pm \textbf{3.29}$ & $\textbf{95.27} \pm \textbf{4.01}$ & $\textbf{94.71} \pm \textbf{4.09}$ \\ \cline{2-9} \hline

\multirow{18}{*}{\textbf{PTB}} & \multirow{8}{*}{\textbf{100\%}} & \textbf{TS2vec} & $85.14 \pm 1.66$ & $87.82 \pm 2.21$ & $76.84 \pm 3.99$ & $79.66 \pm 3.63$ & $90.50 \pm 1.59$ & $90.07 \pm 1.73$ \\
 &  & \textbf{TF-C} & $87.50 \pm 2.43$ & $85.50 \pm 3.04$ & $82.68 \pm 4.04$ & $83.77 \pm 3.50$ & $77.59 \pm 19.22$ & $80.62 \pm 15.10$ \\ 
 &  & \textbf{Mixing-up} & $87.61 \pm 1.48$ & $89.56 \pm 2.80$ & $79.30 \pm 1.67$ & $82.56 \pm 2.00$ & $89.29 \pm 1.26$ & $88.94 \pm 1.04$ \\ 
 &  & \textbf{TS-TCC} & $82.24 \pm 3.55$ & $85.28 \pm 5.08$ & $69.46 \pm 5.85$ & $72.08 \pm 6.85$ & $87.60 \pm 2.51$ & $86.26 \pm 3.00$ \\ 
 &  & \textbf{SimCLR} & $84.19 \pm 1.32$ & $85.85 \pm 1.89$ & $73.89 \pm 2.95$ & $76.84 \pm 2.80$ & $85.85 \pm 1.89$ & $70.70 \pm 2.36$ \\ 
 &  & \textbf{CLOCS} & $86.39 \pm 2.76$ & $87.06 \pm 2.81$ & $77.95 \pm 4.79$ & $80.71 \pm 4.78$ & $90.41 \pm 2.07$ & $87.35 \pm 3.36$ \\ 
 &  & \textbf{COMET} & $87.84 \pm 1.98$ & $87.67 \pm 1.72$ & $81.14 \pm 3.68$ & $83.45 \pm 3.22$ & $92.95 \pm 1.56$ & $87.47 \pm 2.82$ \\
 % &  & \textbf{LMCF*} & $ \textbf{91.33}\pm \textbf{2.03}$ & $\textbf{90.89} \pm \textbf{2.83}$ & $\textbf{84.20} \pm \textbf{4.09}$ & $\textbf{87.64} \pm \textbf{3.82}$ & $\textbf{94.16} \pm \textbf{2.27}$ & $\textbf{90.02} \pm \textbf{3.05}$ \\
 %  &  & \textbf{COMET+AE-GAN*} & $ \textbf{88.49}\pm \textbf{2.06}$ & $\textbf{88.91} \pm \textbf{2.37}$ & $\textbf{83.49} \pm \textbf{4.59}$ & $\textbf{85.67} \pm \textbf{4.07}$ & $\textbf{93.88} \pm \textbf{1.94}$ & $\textbf{88.61} \pm \textbf{3.20}$ \\ 
 &  & \textbf{LMCRD*} & $ \textbf{93.65}\pm \textbf{2.35}$ & $\textbf{93.28} \pm \textbf{1.86}$ & $\textbf{85.39} \pm \textbf{3.84}$ & $\textbf{85.73} \pm \textbf{3.27}$ & $\textbf{95.56} \pm \textbf{1.57}$ & $\textbf{90.28} \pm \textbf{3.49}$ \\ \cline{2-9} 
 & \multirow{8}{*}{\textbf{10\%}} & \textbf{TS2vec} & $82.49 \pm 4.71$ & $80.39 \pm 5.04$ & $83.35 \pm 0.91$ & $80.18 \pm 4.04$ & $93.03 \pm 1.03$ & $92.81 \pm 0.97$ \\
 &  & \textbf{TF-C} & $85.37 \pm 1.23$ & $82.80 \pm 2.35$ & $79.94 \pm 0.71$ & $81.09 \pm 1.14$ & $81.57 \pm 15.60$ & $84.57 \pm 8.12$ \\ 
 &  & \textbf{Mixing-up} & $87.05 \pm 1.41$ & $86.56 \pm 3.24$ & $80.61 \pm 2.68$ & $82.62 \pm 1.99$ & $89.28 \pm 1.43$ & $87.22 \pm 2.76$ \\ 
 &  & \textbf{TS-TCC} & $83.38 \pm 1.53$ & $85.11 \pm 2.11$ & $72.24 \pm 2.45$ & $75.27 \pm 2.64$ & $86.06 \pm 1.76$ & $84.34 \pm 2.08$ \\ 
 &  & \textbf{SimCLR} & $83.84 \pm 2.15$ & $87.19 \pm 1.34$ & $72.51 \pm 4.63$ & $75.44 \pm 4.77$ & $87.19 \pm 1.34$ & $69.99 \pm 3.84$ \\ 
 &  & \textbf{CLOCS} & $88.25 \pm 2.48$ & $88.64 \pm 2.12$ & $81.40 \pm 4.64$ & $83.84 \pm 4.03$ & $91.91 \pm 2.40$ & $89.76 \pm 3.94$ \\ 
 &  & \textbf{COMET} & $88.49 \pm 3.28$ & $88.98 \pm 2.60$ & $81.65 \pm 6.00$ & $84.01 \pm 5.61$ & $94.83 \pm 1.08$ & $92.48 \pm 2.22$ \\
 % &  & \textbf{LMCRF-LMC} & $ \textbf{89.67}\pm \textbf{2.93}$ & $\textbf{89.66} \pm \textbf{2.47}$ & $\textbf{83.32} \pm \textbf{3.62}$ & $\textbf{85.35} \pm \textbf{3.81}$ & $\textbf{95.35} \pm \textbf{3.67}$ & $\textbf{94.94} \pm \textbf{3.71}$ \\
 % &  & \textbf{LMCRD-RD} & $ \textbf{90.84}\pm \textbf{3.01}$ & $\textbf{89.67} \pm \textbf{2.88}$ & $\textbf{83.05} \pm \textbf{2.89}$ & $\textbf{84.68} \pm \textbf{2.95}$ & $\textbf{95.20} \pm \textbf{2.62}$ & $\textbf{94.32} \pm \textbf{3.89}$ \\
 &  & \textbf{LMCRD*} & $ \textbf{91.35}\pm \textbf{2.77}$ & $\textbf{91.56} \pm \textbf{3.12}$ & $\textbf{85.23} \pm \textbf{2.65}$ & $\textbf{85.61} \pm \textbf{3.56}$ & $\textbf{96.30} \pm \textbf{3.22}$ & $\textbf{97.87} \pm \textbf{4.02}$ \\ \cline{2-9} \hline
 
\end{tabular}
}
\end{table*}

\begin{figure}[htbp]
    \centering
    \includegraphics[width=0.5\textwidth]{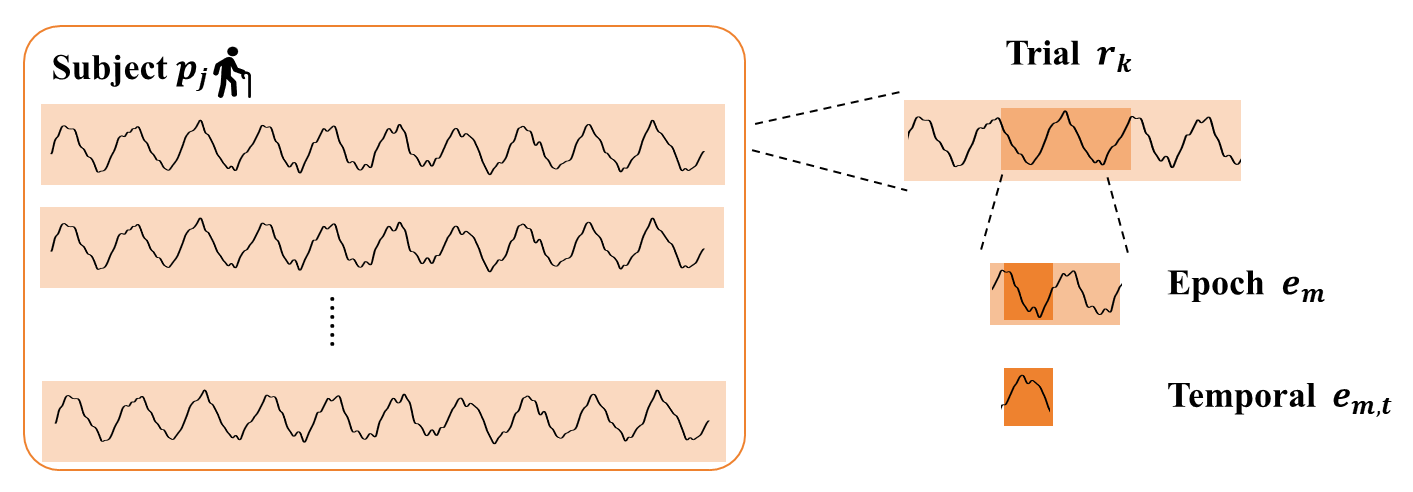} % 替换为图片文件名
    \caption{\textbf{Structure of Medical Time Series.}The data is organized into four levels: Subject (\( p_j \)), Trial (\( r_k \)), Epoch (\( e_m \)), and Temporal (\( e_{m,t} \)), capturing different granularities of the data, from subject-level recordings to temporal observations within individual epochs.}
    \label{fig:sample_image}
\end{figure}

\begin{figure}[htbp]
    \centering
    \includegraphics[width=0.5\textwidth]{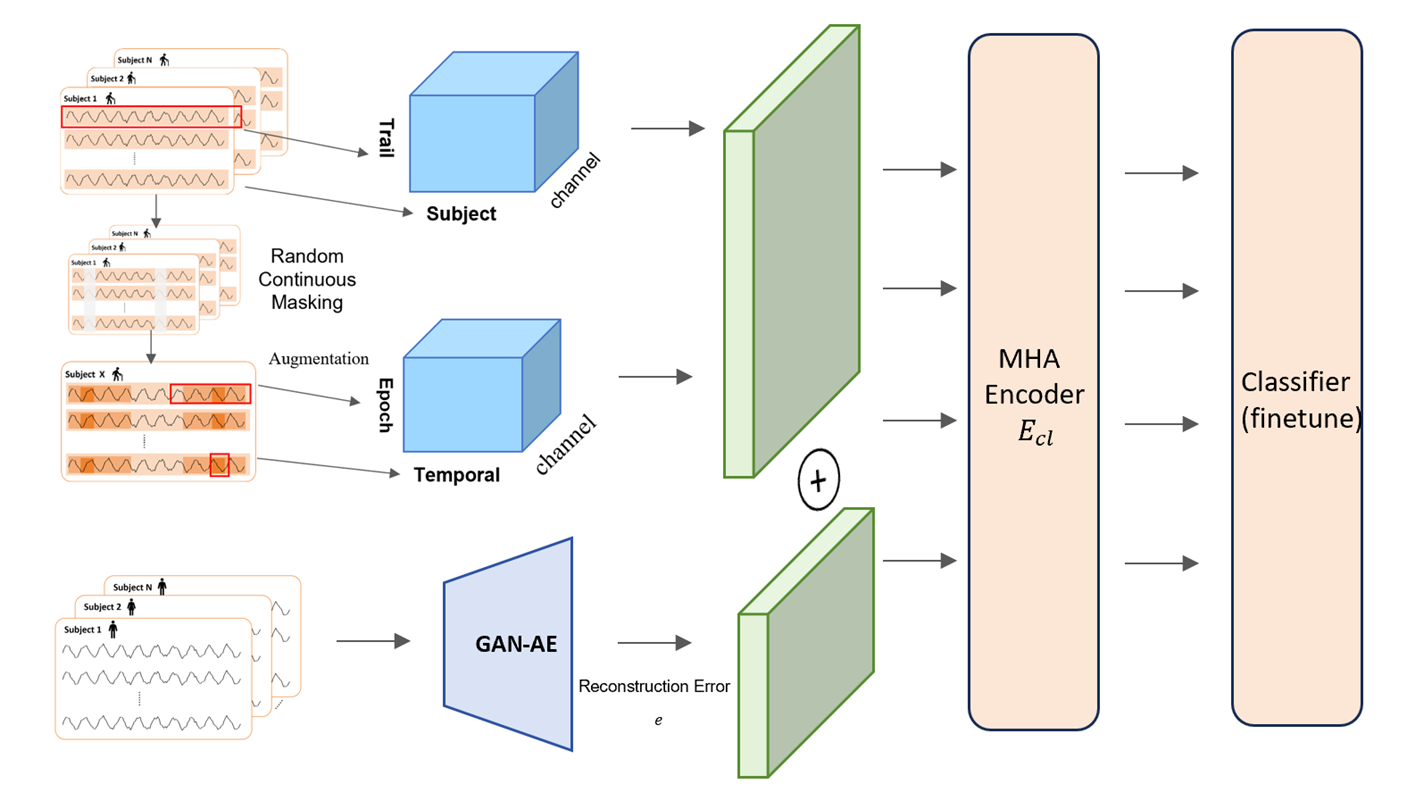} % 替换为图片文件名
    \caption{Overview of LMCRD}
    \label{fig:overview}
\end{figure}

\begin{figure}[htbp]
    \centering
    \includegraphics[width=0.5\textwidth]{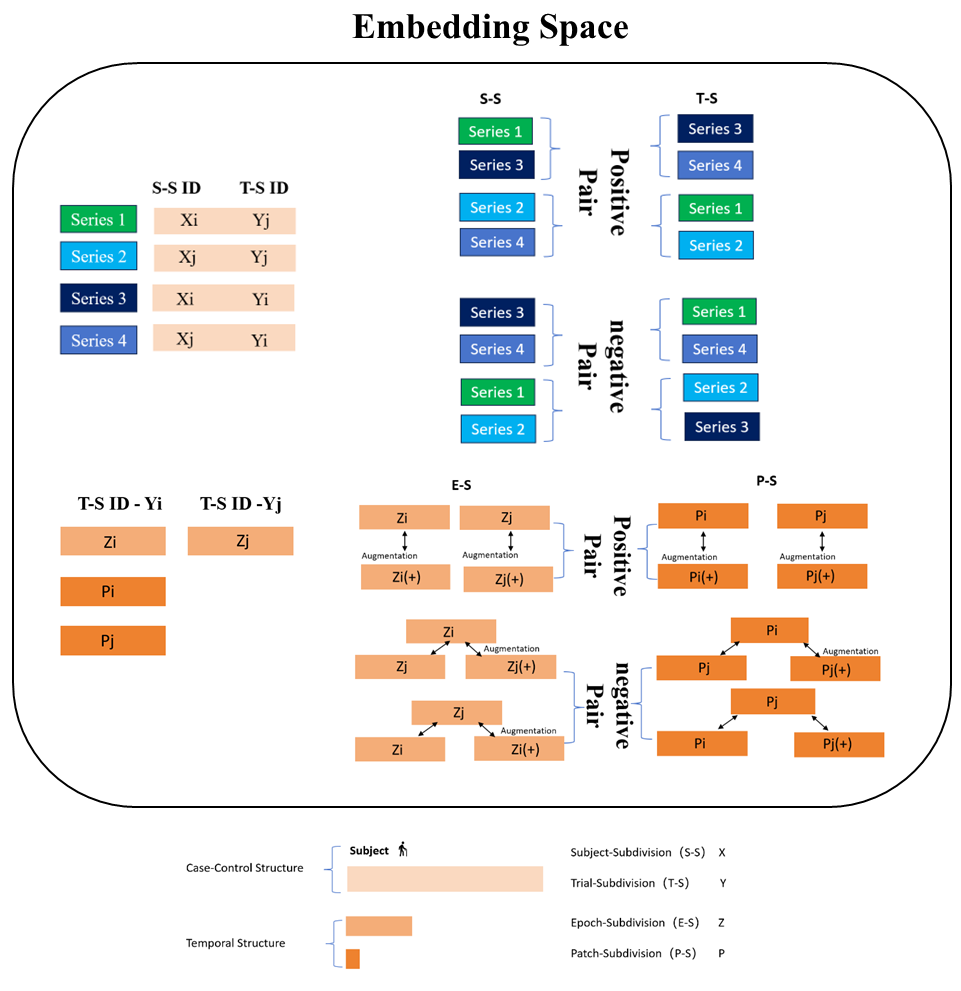} % 替换为图片文件名
    \caption{Embedding Space}
    \label{fig:embedding_space}
\end{figure}

\begin{table*}[t]
\centering
\caption{Ablation Study.}
\label{tab:Ablation Study.}
\resizebox{0.7\textwidth}{!}{% 调整表格宽度以适应页面宽度
\begin{tabular}{c|clllllll}
\hline
\textbf{Dataset} & \textbf{Experiment} & \textbf{Models} & \textbf{Accuracy} & \textbf{Precision} & \textbf{Recall} & F1 \textbf{score} & \textbf{AUROC} & \textbf{AUPRC} \\ \hline
% PFT
\multirow{12}{*}{\textbf{AD}} & \multirow{4}{*}{\textbf{PFT}} & 
\textbf{LMCRD} & $\textbf{80.48}\pm\textbf{5.97}$ & $\textbf{80.34}\pm\textbf{6.00}$ & $\textbf{80.04}\pm\textbf{6.27}$ & $\textbf{80.12}\pm\textbf{6.22}$ & $\textbf{85.96}\pm\textbf{7.17}$ & $8\textbf{5.40}\pm\textbf{7.52}$ \\
& & L\textbf{MCRD-LMC} & $77.42\pm4.40$ & $78.64\pm3.73$ & $76.11\pm4.98$ & $76.26\pm5.26$ & $82.44\pm5.19$ & $81.30\pm5.56$ \\
& & \textbf{LMCRD-RD} & $79.60\pm6.01$ & $80.41\pm5.86$ & $78.50\pm6.51$ & $78.75\pm6.46$ & $86.32\pm6.54$ & $85.85\pm6.99$ \\
& & \textbf{LMCRD-0} & $76.09\pm4.21$ & $77.36\pm3.97$ & $74.68\pm4.62$ & $74.80\pm4.83$ & $81.30\pm4.97$ & $80.50\pm5.31$ \\
% 100
& \multirow{4}{*}{\textbf{FFT 100}\%} &
\textbf{LMCRD} & $\textbf{80.48}\pm\textbf{5.97}$ & $\textbf{80.34}\pm\textbf{6.00}$ & $\textbf{80.04}\pm\textbf{6.27}$ & $\textbf{80.12}\pm\textbf{6.22}$ & $\textbf{85.96}\pm\textbf{7.17}$ & $\textbf{85.40}\pm\textbf{7.52}$ \\
& & \textbf{LMCRD-LMC} & $77.42\pm4.40$ & %
$78.64\pm3.73$ & $76.11\pm4.98$ & $76.26\pm5.26$ & $82.44\pm5.19$ & $81.30\pm5.56$ \\
& & \textbf{LMCRD-R}D & $79.60\pm6.01$ & $80.41\pm5.86$ & $78.50\pm6.51$ & $78.75\pm6.46$ & $86.32\pm6.54$ & $85.85\pm6.99$ \\
& & \textbf{LMCRD-0} & $84.50\pm4.46$ & $88.31\pm2.42$ & $82.95\pm5.39$ & $83.33\pm5.15$ & $94.44\pm2.37$ & $94.43\pm2.48$ \\
% 10
& \multirow{4}{*}{\textbf{FFT 10}\%} &
\textbf{LMCRD} & $\textbf{94.67}\pm\textbf{1.84}$ & $\textbf{94.93}\pm\textbf{1.76}$ & $\textbf{94.41}\pm\textbf{1.99}$ & $\textbf{94.58}\pm\textbf{1.89}$ & $\textbf{98.10}\pm\textbf{1.20}$ & $\textbf{98.18}\pm\textbf{1.15}$ \\
& & \textbf{LMCRD-LMC} & $92.77\pm2.35$ & %
$92.92\pm2.43$ & $92.55\pm2.39$ & $92.66\pm2.37$ & $97.30\pm1.40$ & $97.39\pm1.40$ \\
& & \textbf{LMCRD-RD} & $93.47\pm3.52$ & $93.54\pm3.41$ & $93.68\pm3.16$ & $94.43\pm3.50$ & $87.27\pm1.34$ & $98.30\pm1.34$ \\
& & \textbf{LMCRD-0} & $91.43\pm3.12$ & $92.52\pm2.36$ & $90.71\pm3.56$ & $91.14\pm3.31$ & $96.44\pm2.84$ & $96.48\pm2.82$ \\
\hline
\end{tabular}
}
\end{table*}

\label{sec:appendix_c}
\section{Datasets and Data Preprocessing}

\renewcommand{\thesection}{C\arabic{section}} % 设置主编号为 B1, B2 等
\renewcommand{\thesubsection}{C\arabic{section}.\arabic{subsection}} % 设置子编号为 B1.1, B1.2 等
\setcounter{section}{0} % 重置章节计数器

\section{External Datasets} 
\subsection{Alzheimer’s Disease Dataset}
The AD dataset \cite{miltiadous2023dataset}is a public EEG time series dataset with 3 classes, comprising 36 Alzheimer's disease (AD) patients, 23 Frontotemporal Dementia (FTD) patients, and 29 healthy control (HC) subjects. The dataset contains 19 channels, and the raw sampling rate is 500Hz. Each subject contributes a single trial, with durations of approximately 13.5 minutes for AD subjects (min=5.1, max=21.3), 12 minutes for FTD subjects (min=7.9, max=16.9), and 13.8 minutes for HC subjects (min=12.5, max=16.5). A bandpass filter between 0.5-45Hz is applied to each trial. We preprocess the data by downsampling each trial to 256Hz and segmenting it into non-overlapping 1-second samples with 256 timestamps, discarding any samples shorter than 1 second. From the original 19 channels, we extract 16 overlapping channels for analysis. This preprocessing results in a total of 69,752 samples. For training, validation, and test set splits, we adopt both subject-dependent and subject-independent setups. In the subject-dependent setup, 60\%, 20\%, and 20\% of the total samples are allocated to the training, validation, and test sets, respectively. In the subject-independent setup, 60\%, 20\%, and 20\% of the total subjects, along with their corresponding samples, are allocated to the training, validation, and test sets.
\subsection{Parkinson's Disease Dataset} 
This study utilizes an EEG dataset of 46 participants from a study conducted at the University of New Mexico (UNM, Albuquerque, New Mexico), previously described in \cite{anjum2020linear}. The dataset includes 22 participants with Parkinson’s disease (PD) in an OFF medication state and 24 healthy controls. EEG recordings were obtained using a 64-channel Brain Vision system at a sampling rate of 500 Hz. For our analysis, we selected 33 channels with consistent naming that match the configuration of other datasets. The EEG data were high-pass filtered at 1 Hz to remove noise, and no additional preprocessing steps were applied. Only the first one minute of eyes-closed resting-state EEG data was used for each participant.To process the data, each 1-minute recording was segmented into non-overlapping 5-second samples, resulting in data represented as 
\(\mathbb{R}^{n \times 33 \times 1250}\), where \(n\) is the number of 5-second segments and 1250 corresponds to the number of timestamps per segment.
\subsection{Myocardial Infarction Dataset} 
PTB-XL\cite{wagner2020ptb} is a large-scale public ECG time series dataset containing recordings from 18,869 subjects across 12 channels, with 5 labels representing 4 heart disease categories and 1 healthy control category. Each subject may contribute one or more trials; however, to ensure consistency, we exclude subjects with varying diagnostic results across trials, resulting in a final dataset of 17,596 subjects. The raw trials are recorded as 10-second intervals, available in two versions with sampling frequencies of 100Hz and 500Hz. For this study, we utilize the 500Hz version, which is subsequently downsampled to 250Hz and normalized using standard scalers. Each trial is then segmented into non-overlapping 1-second samples with 250 timestamps, and any samples shorter than 1 second are discarded. This preprocessing results in a total of 191,400 samples. For training, validation, and test set splits, we adopt a subject-independent setup, allocating 60\%, 20\%, and 20\% of the total subjects and their corresponding samples to the training, validation, and test sets, respectively.
\section{Target Datasets}
\subsection{Alzheimer’s Disease Dataset}
The AD dataset\cite{escudero2006analysis} consists of EEG recordings from 22 subjects, including 11 patients with probable Alzheimer's disease (5 men, 6 women; age: 72.5 ± 8.3 years, mean ± standard deviation) and 11 healthy controls. The patients were recruited from the Alzheimer’s Patients’ Relatives Association of Valladolid (AFAVA) and underwent thorough clinical evaluations, including neurological and physical examinations, brain scans, and MMSE assessments to evaluate cognitive ability. EEG recordings were conducted at the University Hospital of Valladolid (Spain) using Profile Study Room 2.3.411 EEG equipment (Oxford Instruments) at 19 electrode positions (F3, F4, F7, F8, Fp1, Fp2, T3, T4, T5, T6, C3, C4, P3, P4, O1, O2, Fz, Cz, Pz) based on the international 10-20 system. Each subject contributed over 5 minutes of EEG data recorded in a relaxed state with eyes closed and awake to minimize artifacts. The recordings were sampled at 256 Hz with 12-bit A-to-D precision.
Each patient contributes an average of 30.0 ± 12.5 trials, with each trial spanning a 5-second interval and comprising 16 channels. Prior to analysis, trials were standardized using a standard scaler and segmented into nine half-overlapping samples, each containing 256 timestamps (equivalent to 1 second). The dataset includes a total of 5967 multivariate EEG samples, divided into 4329 training samples, 891 validation samples, and 747 test samples. A patient-independent split was used, with patient IDs 17 and 18 allocated to the validation set, IDs 19 and 20 to the test set, and the remaining samples to the training set. Each sample is labeled with a binary value indicating the presence of Alzheimer’s disease and is assigned unique trial and patient identifiers to preserve its origin.
\subsection{Parkinson's Disease Dataset} 
The TDBrain dataset\cite{van2022two}contains brain signal recordings from 1274 patients across 33 channels at 500 Hz during Eye Closed (EC) and Eye Open (EO) tasks, covering 60 different diseases, with some patients diagnosed with multiple conditions. This study focuses on a subset of 25 Parkinson’s disease patients and 25 healthy controls, utilizing only the EC task for representation learning.
Raw EC trials are preprocessed using a sliding window approach that divides each trial into pseudo-trials of 2560 timestamps (10 seconds) after resampling to 256 Hz. These pseudo-trials are standardized and further split into 19 half-overlapping samples, each consisting of 256 timestamps (1 second). Each sample is assigned a binary label (Parkinson’s or healthy), along with patient and trial identifiers, where trial IDs correspond to the processed pseudo-trials.
The data is split in a patient-independent manner, with samples from 8 patients (4 healthy and 4 Parkinson’s) forming the validation set, another 8 patients forming the test set, and the remaining patients used for training.
\subsection{Myocardial Infarction Dataset}
The PTB dataset\cite{goldbergerphysionet}contains ECG recordings from 290 patients, spanning 15 channels sampled at 1000 Hz, and includes eight types of heart diseases. This study focuses on a subset with 198 patients for binary classification, specifically distinguishing between Myocardial Infarction and healthy controls. The ECG signals are resampled to 250 Hz, normalized using a standard scaler, and segmented into individual heartbeats to preserve crucial peak information, as a standard sliding window approach may lead to information loss.The segmentation process involves several steps: (1) determining the maximum heartbeat duration by calculating the median R-peak interval across all channels in each trial, excluding outliers; (2) identifying the position of the first R-peak using the median value across channels; (3) splitting raw trials into individual heartbeats based on the median R-peak interval, with segments extending symmetrically from the R-peak; and (4) applying zero-padding to ensure uniform sample lengths.The dataset is split in a patient-independent manner, with samples from 28 patients (7 healthy and 21 Myocardial Infarction) assigned to both the validation and test sets, while the remaining samples are used for training.

% 切换到 D 开头的编号
% \renewcommand{\thesection}{D\arabic{section}} % 设置主编号为 D1, D2 等
\renewcommand{\thesection}{D} % 设置主编号为 D1, D2 等
\setcounter{section}{0} % 重置章节计数器
\label{sec:appendix_d}
\section{Loss Functions}

The model is trained by optimizing two loss functions: the discriminator loss and the generator loss. During the training process, the discriminator is first updated using the discriminator loss ($L_D$), followed by the generator. The two components are updated alternately.

- \textbf{Discriminator Loss}:
  \[
  L_D = \frac{1}{2} (L_{D_{\text{real}}} + L_{D_{\text{fake}}})
  \]
  where:
  - $L_{D_{\text{real}}} = \text{BCE}(C(xx_i), 1)$: The loss for real data, encouraging the discriminator to classify real data as 1.
  - $L_{D_{\text{fake}}} = \text{BCE}(C(\hat{xx}_i), 0)$: The loss for generated data, encouraging the discriminator to classify generated data as 0.
  - $\text{BCE}$ is the binary cross-entropy loss function.

- \textbf{Generator Loss}:
  \[
  L_G = L_{G_{\text{adv}}} + L_{G_{\text{recon}}}
  \]
  where:
  - $L_{G_{\text{adv}}} = \text{BCE}(C(\hat{xx}_i), 1)$: The adversarial loss, encouraging the discriminator to classify generated data as 1.
  - $L_{G_{\text{recon}}} = \text{MSE}(\hat{xx}_i, xx_i)$: The reconstruction loss, ensuring that the generated data $\hat{xx}_i$ is close to the original data $xx_i$.
  - $\text{MSE}$ is the mean squared error loss function.

% 切换到 E 开头的编号
% \renewcommand{\thesection}{D\arabic{section}} % 设置主编号为 E1, E2 等
\renewcommand{\thesection}{E} % 设置主编号为 D1, D2 等
\setcounter{section}{0} % 重置章节计数器
\label{sec:appendix_d}
\section{Ablation Study}
To validate the effectiveness of the AE-GAN knowledge transfer module and the Learnable Multi-Views Contrastive Framework (LMCF) in the LMCRD model, we conduct an ablation study on the AD dataset. Specifically, we compare the full LMCRD model with three ablated variants across three evaluation tasks: \textbf{PFT}, \textbf{FFT100\%}, and \textbf{FFT10\%}.According to \ref{tab:Ablation Study.}, the three variants include LMCRD-LMC, which removes the AE-GAN-based knowledge transfer module while retaining LMCF, leading to an accuracy drop to \textbf{77.42 ± 4.40\%} and an F1 score reduction to \textbf{76.26 ± 5.26\%}; LMCRD-RD, which removes LMCF while preserving the AE-GAN-based knowledge transfer module, resulting in an accuracy decrease to \textbf{79.60 ± 6.01\%} and an F1 score drop to \textbf{78.75 ± 6.46\%}; and LMCRD-0, which serves as the baseline model by removing both modules, yielding the lowest accuracy of \textbf{77.42 ± 4.40\%} and an F1 score of \textbf{76.26 ± 5.26\%}. The experimental results, summarized in \textbf{Table}~\ref{tab:Ablation Study.}, show that compared to the full LMCRD model with an accuracy of \textbf{80.48 ± 5.97\%} and an F1 score of \textbf{80.12 ± 6.22\%}, LMCRD-LMC, LMCRD-RD, and LMCRD-0 experience accuracy drops of \textbf{3.06\%}, \textbf{0.88\%}, and \textbf{3.06\%}, respectively, along with corresponding F1 score reductions of \textbf{3.86\%}, \textbf{1.37\%}, and \textbf{3.86\%}. These results highlight that the AE-GAN module effectively transfers knowledge from external datasets to the target dataset, enriching the probability distribution of normal samples and improving classification performance by mitigating data scarcity and distribution shifts. The significant performance degradation observed upon removing this module underscores its crucial role in enhancing the model’s generalization capability. Furthermore, LMCF leverages a multi-head attention mechanism to model multi-view representations of the data, capturing different perspectives of the same sample. By integrating contrastive learning, LMCF enables adaptive learning of positive and negative sample pairs, thereby improving the model's ability to differentiate between subtle variations. The performance drop after removing LMCF suggests that it plays a key role in refining decision boundaries and capturing multi-view structures within the data. In summary, both the AE-GAN knowledge transfer module and the Learnable Multi-Views Contrastive Framework are indispensable in LMCRD. Their synergy not only enhances the model’s ability to distinguish normal and abnormal samples but also strengthens its capability to learn multi-view representations, ultimately leading to superior classification performance.

\end{document}